\documentclass[10pt,twocolumn,letterpaper]{article}
\usepackage[accsupp]{axessibility}  %
\usepackage{iccv}
\makeatletter
\@namedef{ver@everyshi.sty}{}
\makeatother

\usepackage{times}
\usepackage{epsfig}
\usepackage{graphicx}
\usepackage{amsmath}
\usepackage{amssymb}
\usepackage{capt-of}
\usepackage[caption=false]{subfig}
\usepackage[table]{xcolor}
\usepackage{tikz}
\usepackage{bm}
\definecolor{yellow}{rgb}{1,1, 0.6}
\definecolor{lightyellow}{rgb}{1,1, 0.8}
\definecolor{orange}{rgb}{1, 0.8, 0.6}
\definecolor{tab_red}{rgb}{1, 0.6, 0.6}

\newcommand\ours{Tetra-NeRF\xspace}
\newcommand\suppmat{\textit{Supp. Mat.}\xspace}

\usetikzlibrary{positioning}
\usetikzlibrary{calc}
\usepackage[bottom]{footmisc}
\usepackage{cite}

\usepackage{enumitem}

\newcommand{\PAR}[1]{\vskip4pt \noindent{\bf #1~}}

\usepackage[pagebackref=true,breaklinks=true,letterpaper=true,colorlinks,bookmarks=false]{hyperref}

\definecolor{mycitecolor}{HTML}{195a66}
\hypersetup{
  citecolor  = mycitecolor,
  colorlinks = true,
}

\iccvfinalcopy
\def\iccvPaperID{4970}

\begin{document}

\title{Tetra-NeRF: Representing Neural Radiance Fields Using Tetrahedra}

\author{Jonas Kulhanek\\
Czech Technical University in Prague\\
{\tt\small jonas.kulhanek@cvut.cz}
\and
Torsten Sattler\\
Czech Technical University in Prague\\
{\tt\small torsten.sattler@cvut.cz}
}

\twocolumn[{
\maketitle

\vspace{-1.2em}
\vspace{-0.5em}
\centering
\includegraphics[width=0.8\textwidth]{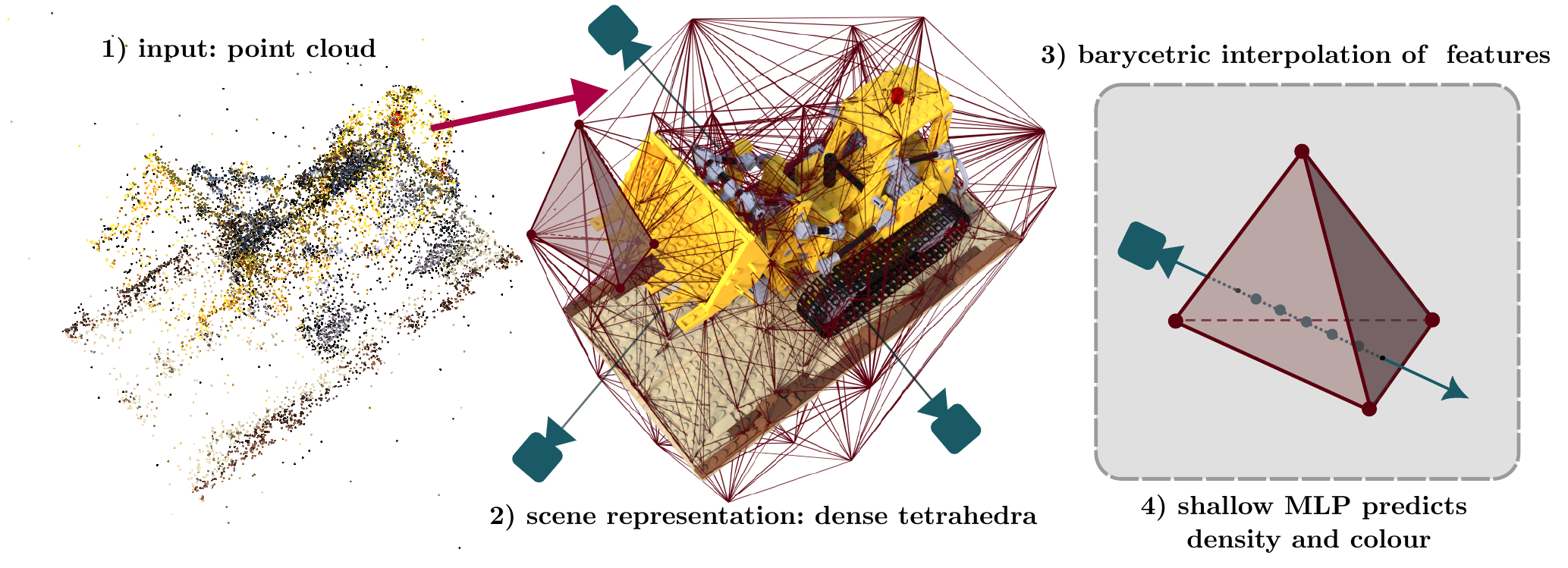}
\vspace{-0.5em}
\captionof{figure}{
The input to \ours is a point cloud which is triangulated to get a set of tetrahedra used to represent the radiance field. Rays are sampled, and the field is queried. Barycentric interpolation is used to interpolate tetrahedra vertices, and the resulting features are passed through a shallow MLP to get the density and colours for volumetric rendering.
\label{fig:teaser}
}
\vspace{1.4em}
}]

\begin{abstract}
Neural Radiance Fields (NeRFs) are a very recent and very popular approach for the problems of novel view synthesis and 3D reconstruction. A popular scene representation used by NeRFs is to combine a uniform, voxel-based subdivision of the scene with an MLP. Based on the observation that a (sparse) point cloud of the scene is often available, this paper proposes to use an adaptive representation based on tetrahedra obtained by  Delaunay triangulation instead of uniform subdivision or point-based representations. We show that such a representation enables efficient training and leads to state-of-the-art results. Our approach elegantly combines concepts from 3D geometry processing, triangle-based rendering, and modern neural radiance fields. Compared to voxel-based representations, ours provides more detail around parts of the scene likely to be close to the surface. Compared to point-based representations, our approach achieves better performance. 
The source code is 
publicly 
available at: \href{https://jkulhanek.com/tetra-nerf/}{https://jkulhanek.com/tetra-nerf}.
\end{abstract}

\section{Introduction}
Reconstructing 3D scenes from images and rendering photo-realistic novel views is a key problem in computer vision.
Recently, NeRFs \cite{mildenhall2021nerf,barron2021mipnerf,barron2022mipnerf360} became dominant in the field for their superior photo-realistic results.
Originally, NeRFs used MLPs to represent the 3D scene as an implicit function.
Given a set of posed images, NeRF randomly samples a batch of pixels, casts rays from the pixels into the 3D space, queries the implicit function at randomly sampled distances along the rays, and aggregates the sampled values using volumetric rendering \cite{mildenhall2021nerf,max1995optical}. While the visual results of such methods are of high quality, the problem is that querying large MLPs at millions of points is costly. Also, once the network is trained, it is difficult to make any changes to the represented radiance field as everything is baked into the MLPs parameters, and any change has a non-local effect.
Since then, there have been a lot of proposed alternatives to the large MLP field representation \cite{muller2022ingp,chen2022tensorf,fridovich2022plenoxels,yu2021plenoctrees,xu2022pointnerf,sun2022dvgo,liu2020nsvf}. These methods %
combine %
an MLP with a voxel feature grid \cite{muller2022ingp,sun2022dvgo}, or in some cases represent the radiance field directly as a tensor \cite{chen2022tensorf,chen2023factorfields,fridovich2022plenoxels}.
When querying these representations, first, the containing voxel is found, and the features stored at the eight corner points of the voxel are trilinearly interpolated. The result is either passed through a shallow MLP \cite{muller2022ingp,sun2022dvgo,chen2023factorfields,chen2022tensorf} or is used directly as the density and colour \cite{fridovich2022plenoxels,sun2022dvgo,liu2020nsvf}.

Having a dense tensor represent the entire scene is very inefficient, as we only need to represent a small space around surfaces. Therefore, different methods propose different ways of tackling the issue. Instant-NGP \cite{muller2022ingp}, for example, uses a hash grid instead of a dense tensor, where it relies on optimisation to resolve the hash collisions. However, similarly to MLPs, any change to the stored hashmap influences the field in many places. A more common direction to addressing the issue is by directly using a sparse tensor representation \cite{fridovich2022plenoxels,chen2022tensorf}. These methods start with a low-resolution grid and, at predefined steps, subsample the representation, increasing the resolution.
These approaches tend to require a careful setting of hyperparameters, such as the scene bounding box and the subdivision steps, in order for the methods to work well.

Because many of these methods use traditional structure from motion (SfM) \cite{schoenberger2016sfm,schoenberger2016mvs} methods to generate the initial poses for the captured images, we can reuse the original reconstruction in the scene representation.
Inspired by classical surface reconstruction methods \cite{hiep2009towards,labatut2009robust,labatut2007efficient,jancosek2011multi}, we %
represent the scene as a dense triangulation of the input point cloud, where the scene is a set of non-overlapping tetrahedra whose union is the convex hull of the original point cloud \cite{delaunay1934bulletin}. When querying such a representation, we find to which tetrahedron the query point belongs and perform barycentric linear interpolation of the features stored in the vertices of the tetrahedron. This very simple representation can be thought of as the direct extension of the classical triangle-rendering pipelines used in graphics \cite{parker2010optix,phong1975illumination,moller1997fast}.
The representation avoids problems with %
the sparsity of the input point cloud as the tetrahedra fully cover the scene, resulting in a continuous rather than discrete representation.

This paper makes the following contributions: 
(\textbf{1}) We propose a novel radiance field representation which is initialised from a sparse or dense %
point cloud. This representation is naturally denser in the proximity of surfaces and, therefore, provides a higher resolution in these regions. 
(\textbf{2}) The proposed representation is evaluated on multiple synthetic and real-world datasets and is compared with a state-of-the-art point-cloud-based representation -- Point-NeRF \cite{xu2022pointnerf}. 
The presented %
results show that our method clearly outperforms this baseline. 
We further demonstrate the effectiveness of our adaptive representation by comparing it with a voxel-based representation that uses the same number of trainable parameters. %
(\textbf{3}) We make the source code and model checkpoints publicly available.\footnote{\url{https://github.com/jkulhanek/tetra-nerf}}

\section{Related work}
\PAR{Multi-view reconstruction.}
The problem of multi-view reconstruction has been studied extensively and tackled with a variety of structure from motion (SfM) \cite{tang2018banet,schoenberger2016sfm,vijayanarasimhan2017sfm}, and multi-view stereo (MVS) \cite{schoenberger2016mvs,furukawa2009accurate,cheng2020deep,yao2018mvsnet} methods.
These methods usually output the scene represented as a point cloud \cite{schoenberger2016sfm,schoenberger2016mvs}.
In most rendering approaches, the point cloud is converted into a mesh \cite{lorensen1987marching,kazhdan2006poisson}, and novel views are rendered by reprojecting observed images into each novel viewpoint and blending them together
using either heuristically-defined \cite{buehler2001unstructured,debevec1996modeling,wood2000surface} or learned \cite{hedman2018deep,riegler2020free,riegler2021stable,yariv2023bakedsdf} blending weights.

However, the process of getting the meshes is usually quite noisy, and the resulting meshes tend to have inaccurate geometry in regions with fine details or complex materials.
Instead of using noisy meshes, point-based neural rendering methods \cite{kopanas2021point,ruckert2022adop,meshry2019neural,aliev2020neural} perform splatting of neural features and use 2D convolutions to render them.
In contrast to these methods, our approach operates and aggregates features directly in 3D and does not suffer from the noise in the point cloud or the reconstructed mesh.

\PAR{Neural radiance fields.}
Recently, NeRFs \cite{mildenhall2021nerf,zhang2020nerf++,barron2021mipnerf,barron2022mipnerf360,lombardi2019neural} have gained a lot of attention thanks to their high-quality rendering performance. The original NeRF method \cite{mildenhall2021nerf} was extended to better handle aliasing artefacts in \cite{barron2021mipnerf}, to better represent unbounded scenes in \cite{zhang2020nerf++,barron2022mipnerf360,reiser2023merf}, or to handle real-world captured images \cite{martin2021nerfw,tancik2022blocknerf}.
The training of the large MLPs used in these methods can be quite slow, and there has been a lot of effort on speeding up either the training \cite{muller2022ingp,fridovich2022plenoxels,chen2022tensorf} or the rendering \cite{reiser2023merf,reiser2021kilonerf,yu2021plenoctrees,hedman2021baking} sometimes at the cost of larger storage requirements.
Other approaches tackled different aspects of NeRFs like view-dependent artefacts \cite{verbin2022ref}, relighting \cite{bi2020neural,boss2021nerd}, or proposed generative models \cite{wang2022clip,poole2022dreamfusion}. Also, a popular research direction is making the models generalize across different scenes \cite{chen2021mvsnerf,reizenstein2021common,kulhanek2022viewformer,yu2021pixelnerf,wang2021ibrnet}.
A large area of research is dedicated to the surface reconstruction and, instead of using the radiance fields, represents the scene implicitly by modelling the signed distance function (SDF) \cite{yu2022monosdf,yariv2021volsdf,wang2021neus,wang2022neus2,rosu2023permutosdf}.
Unlike those approaches, we focus only on the radiance field representation and consider these methods orthogonal to ours.

Although there are some methods that train radiance fields while fine-tuning the poses \cite{lin2021barf,tancik2023nerfstudio} or without known cameras \cite{wang2021nerf--,bian2022nopenerf}, most methods need camera poses for the reconstruction. 
SfM, \eg, COLMAP \cite{schoenberger2016sfm,schoenberger2016mvs}, is typically used for estimating the poses, which also produces a (sparse) point cloud. 
Our approach makes use of this by-product of the pose recovery process instead of only using the poses themselves. %

\PAR{Field representations.}
When a single MLP is used to represent the entire scene, everything is baked into a single set of parameters which cannot be easily modified, because any change to the parameters has a non-local effect, \ie, it changes the scene at multiple unrelated places.
To overcome this problem, others have experimented with different representations of the radiance fields \cite{fridovich2022plenoxels,sun2022dvgo,muller2022ingp,chen2022tensorf,xu2022pointnerf,liu2020nsvf,chen2023factorfields,peng2020convolutional}.
A common practice is to represent the scene as a shallow MLP and an efficiently encoded voxel grid \cite{liu2020nsvf,muller2022ingp,chen2022tensorf,chen2023factorfields,peng2020convolutional}. %
The encoded voxel grid can also be used represent the radiance field directly \cite{fridovich2022plenoxels,sun2022dvgo}.
The voxel grid can be encoded as a sparse tensor \cite{sun2022dvgo,fridovich2022plenoxels,liu2020nsvf}, a factorisation of the 4D tensor \cite{chen2022tensorf,chen2023factorfields}, or a hashmap \cite{muller2022ingp}.
When these structures are queried, trilinear interpolation is used to combine the feature vectors stored in the containing voxel corners.
Unfortunately, the hashmaps \cite{muller2022ingp} and hierarchical representations \cite{chen2023factorfields} have the same non-local effect problem, and the rest of the approaches rely on subsequent upsampling of the field and can be overly complicated. Also, feature vectors cannot be placed arbitrarily in the 3D space as they must lie on the grid with a fixed resolution.
In contrast, our approach is much more flexible as it stores the feature vectors freely in 3D space.

Finally, Point-NeRF \cite{xu2022pointnerf} represents the scene as a point cloud of features which are queried using $k$-nearest neighbours search. However, when the point cloud contains sparse regions, the rays do not intersect any neighbourhoods of any points and the pixels stay empty without the ability to optimise. Therefore, Point-NeRF \cite{xu2022pointnerf} relies on gradually adding more points during training and increasing the scene complexity.
Since we use the triangulation of the point cloud rather than the discrete point cloud itself, our representation is continuous and does not suffer from empty regions. Therefore, we do not have to add any points during training.

\begin{figure}[t!]
    \centering
    \subfloat[\centering dense point cloud]{{
        \includegraphics[width=0.45\columnwidth]{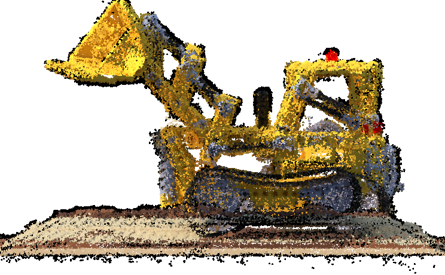}
    }}        
    \subfloat[\centering dense tetrahedra slice]{{
        \includegraphics[width=0.45\columnwidth]{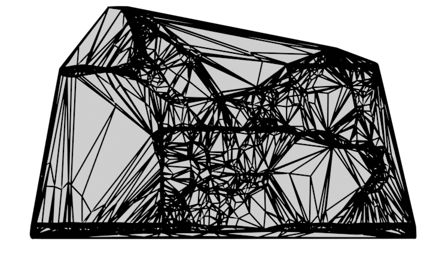}
    }}
    \\
    \subfloat[\centering sparse point cloud]{{
        \includegraphics[width=0.45\columnwidth]{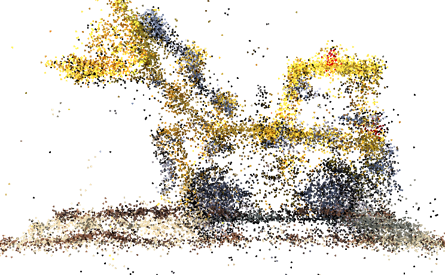}
    }}        
    \subfloat[\centering sparse tetrahedra slice]{{
        \includegraphics[width=0.45\columnwidth]{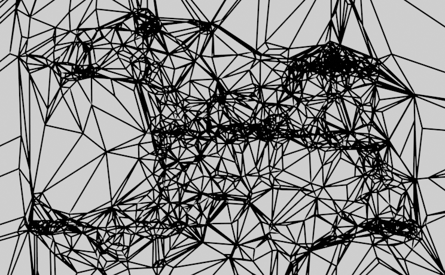}
    }}
    \caption{Input point cloud and a slice through the triangulated tetrahedra. Note that smaller tetrahedra are created closer to the surface of the scene, \ie, regions close to the surface are represented with a finer resolution.}%
    \label{fig:tetrahedra}%
\end{figure}

\section{Method}
A common strategy in the literature is to represent the scene explicitly through a voxel volume. 
In contrast to this uniform subdivision, we investigate using an adaptive subdivision of the scene. 
In many scenarios, an approximation of the scene geometry is either given, \eg, when using SfM to compute the input camera poses, %
or can be computed, \eg, via MVS or single-view depth predictions~\cite{zhao2020monocular}. 
This allows us to compute an adaptive subdivision of the scene via Delaunay triangulation \cite{delaunay1934bulletin} of such a point cloud. 
This results in a set of 
non-overlapping tetrahedra, where smaller tetrahedra are created close to the surface of the scene (\cf Fig.~\ref{fig:tetrahedra}). 
In the following, we explain how this adaptive subdivision of the scene can be used instead of voxels for volume rendering and neural rendering. %

\begin{figure}[t!]
\centering
\quad
\subfloat[\centering trilinear interpolation]{{
    \includegraphics[width=0.45\columnwidth]{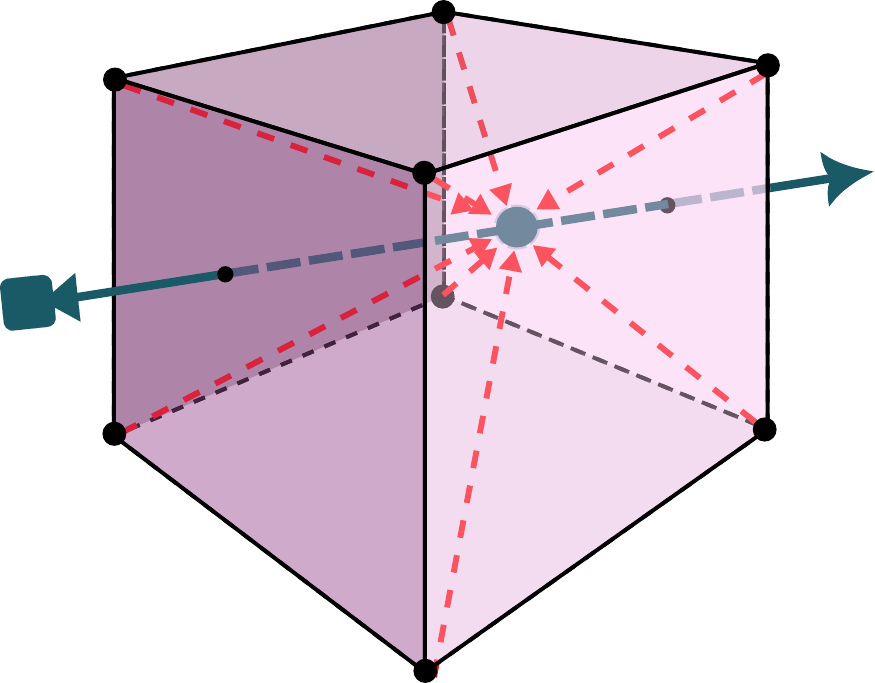}
}}        
\subfloat[\centering barycentric interpolation]{{
\includegraphics[width=0.35\columnwidth]{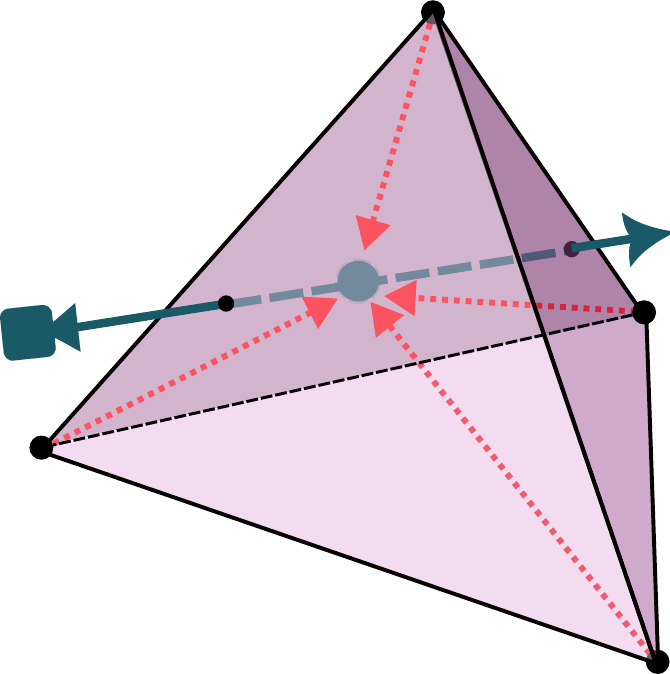}
    \quad
}}
\vspace{0.3\baselineskip}
\caption{\textbf{Trilinear and barycentric intepolation.} Trilinear interpolation is a weighted combination of the eight voxel corners. Barycentric interpolation weights the four vertices of the tetrahedron vertices based on the barycentric coordinates~\cite{floater2015generalized}.
\label{fig:interpolation}
}
\end{figure}

\subsection{Preliminaries}
\par\noindent\textbf{Neural Radiance Fields (NeRFs)~\cite{mildenhall2021nerf}} %
represent a scene through an implicit function $F(\mathbf{x}, \mathbf{d})$, often modelled via a neural network, that returns a colour value and a volume density prediction for a given 3D point $\mathbf{x}$ in the scene observed from a viewing direction $\mathbf{d}$. 
Volume rendering~\cite{max1995optical} is used to synthesise novel views: for each pixel in a virtual view, we project a ray from the camera plane into the scene and sample the radiance field to obtain the colour and density values $\mathbf{c}_i$ and $\sigma_i$ at distances $t_i$ along the ray, where $i=1 \ldots N$. %
The individual samples are then combined to predict the colour $\mathbf{C}$ for the pixel: 
\begin{align}\label{eq:volumetric-rendering}
\begin{split}
\mathbf{C} &= \sum_{i=1}^{N} T_i (1 - \exp(\sigma_i \delta_i))\mathbf{c}_i \enspace , \text{where} \\
T_i &= \exp(-\sum_{i=1}^{N - 1} \sigma_i \delta_i) \enspace ,
\end{split}
\end{align}
and $\delta_i = t_i - t_{i-1}$ is the distance between adjacent samples. 
This process is fully differentiable. 
Thus, after computing the mean squared error (MSE) between the predicted and ground truth colours associated with each ray, we can back-propagate the gradients to the radiance field.

\par\noindent\textbf{Voxel-based feature fields} such as NSVF \cite{liu2020nsvf} represent the scene as a voxel grid and an MLP. 
Each grid point of the voxel grid is assigned a trainable vector. There are eight feature vectors associated with a single voxel, but the vectors are shared between neighbouring voxels. For each query point sampled along the ray, its corresponding voxel is found first. 
A feature for the point is then computed via trilinear interpolation of the features of the voxel grid points based on the position of the query point (\cf Fig.~\ref{fig:interpolation}, left). 
The resulting feature vector is passed through an MLP to predict the density and appearance vector. The appearance vector is combined with the ray direction and passed through a second MLP in order to compute a view-dependent colour.

\subsection{Tetrahedra fields}
Given a set of points in 3D space, we build the tetrahedra structure by triangulating the points. We apply the Delaunay triangulation \cite{delaunay1934bulletin} to obtain a set of non-overlapping tetrahedra whose union is the convex hull of the original points. Fig.~\ref{fig:tetrahedra} shows example tetrahedra obtained by triangulating dense and sparse COLMAP-reconstructed point clouds \cite{schoenberger2016sfm,schoenberger2016mvs}. %
Note that the resulting representation is adaptive as it uses a higher resolution (smaller tetrahedra) closer to the surface and larger tetrahedra to model regions farther away from the surface. 

We associate all vertices of the tetrahedra with trainable vectors.
As in the voxel grid case, vertices, and thus vectors, are shared between adjacent tetrahedra. %
The resulting tetrahedra field can be queried in the same way a voxel grid representation is queried: for each query point, we first find the tetrahedron containing the point. %
Instead of the trilinear interpolation used for voxel volumes, we use the barycentric interpolation \cite{floater2015generalized} to compute a feature vector for the query point from the four feature vectors stored at the tetrahedron's vertices (\cf Fig.~\ref{fig:interpolation}). 
To this end, we compute the query point $\mathbf{x}$'s barycentric coordinates $\bm{\lambda}$, which express the point's 3D coordinates as a unique weighted combination of the 3D coordinates of the tetrahedron's vertices. 
In particular, the weight for a vertex is the volume of the tetrahedron constructed from the query point and the face opposite to the vertex divided by the volume of the full tetrahedron: 
\begin{equation}
\bm{\lambda} = \bigg(\frac{V_{x234}}{V_{1234}}, \frac{V_{1x34}}{V_{1234}}, \frac{V_{12x4}}{V_{1234}}, \frac{V_{123x}}{V_{1234}}\bigg) \,,
\end{equation}
where $V_{1234}$ is the volume of the full tetrahedron and $V_{x234}$, $V_{1x34}$, \ldots, are volumes of tetrahedra with $1^{\text{st}}$, $2^{\text{nd}}$, \ldots, vertex replaced by $\mathbf{x}$.
The same weights $\bm{\lambda}$ are applied to the feature vectors of the vertices to obtain the query feature. 

The interpolated features are used as the input to a small MLP in order to predict density and colour at the query point. We first pass the barycentric-interpolated features through a three-layer MLP to compute the density and appearance features. We then concatenate the appearance features with the ray direction vector encoded using Fourier features \cite{tancik2020fourier,mildenhall2021nerf} and pass the result through a single linear layer to get the raw RGB colour value.

Finally, to map the raw density values $\bar{\sigma_i}$ returned by the network to the volume density $\sigma_i$ required by volume rendering, %
we apply the softplus activation function \cite{barron2021mipnerf}. %
For the RGB colour value, we use the sigmoid function \cite{fridovich2022plenoxels}.%

\subsection{Efficiently querying a tetrahedra field}
Determining the corresponding voxel for a given query point can be done highly efficiently via hashing~\cite{niessner2013hashing}. 
In contrast, finding the corresponding tetrahedron for a query point is more complex. 
This in turn can significantly impact rendering, and thus training, efficiency~\cite{fridovich2022plenoxels,muller2022ingp,sun2022dvgo}. 

In order to efficiently look up the corresponding tetrahedra, we exploit that we are not considering isolated points, but points sampled from a ray.  
We compute the tetrahedra that are intersected by the ray, allowing us to march through the tetrahedra rather than computing them individually per point. 
The relevant tetrahedra can be found efficiently using acceleration structures for fast ray-triangle intersection computations, \eg, via NVidia's OptiX library~\cite{parker2010optix}: %
We first compute the intersections between the rays from a synthetic view and all faces of the tetrahedra. %
For each ray, we take the first 512 intersected triangles\footnote{For efficiency, we only consider a fixed number of triangles. As discussed later on, this can degrade results in larger scenes / scenes with a fine-grained tetrahedralisation, where more intersections are needed. Naturally, more triangles can be considered at the cost of longer run-times.}, and determine the corresponding tetrahedra. 
The tetrahedra can be ordered based on the intersections along the ray, allowing us to easily march through the tetrahedra to determine which tetrahedron to use for a given query point. 

A side benefit of computing ray-triangle intersections is that we can simplify computing the barycentric coordinates: 
For each intersection, we compute the 2D barycentric coordinates \wrt the triangle. 
We then obtain the 3D barycentric coordinates \wrt the associate tetrahedron by simply adding zero for the vertex opposite to the triangle. 
For a query point inside a tetrahedron, we can compute its tetrahedron barycentric coordinates by linearly interpolating between the barycentric coordinates of the two intersections of the ray and the tetrahedron.

\subsection{Coarse and fine sampling}
We follow the common practice of having a two-stage sampling procedure \cite{mildenhall2021nerf,chen2022tensorf}.  In the coarse stage, we sample uniformly along the ray. 
In the fine stage, we use the density weights from the coarse sampling stage to bias the sampling towards sampling closer to the potential surface. Following \cite{mildenhall2021nerf}, we use the stratified uniform sampling for the coarse stage. The stratified uniform sampling splits the ray into equally long intervals and samples uniformly in each interval. Unlike NeRF \cite{mildenhall2021nerf}, we limit the sampling to the space occupied by tetrahedra. In the fine sampling stage, we use the same network as in the coarse sampling stage. 

For the fine sampling stage, we take the accumulated weights $w_i$ from the coarse sampling:
\begin{equation}
\bar{w_i} = (1 - \exp(-\sigma_i \delta_i))\exp\bigg(-\sum_{j=1}^{i-1}\sigma_j \delta_j\bigg) \enspace .
\end{equation}
These weights are the coefficients used in Equation~\ref{eq:volumetric-rendering} as multipliers for the colours \cite{mildenhall2021nerf}. We obtain $w_i$ by normalizing $\bar{w_i}$. Following \cite{mildenhall2021nerf}, we sample set of fine samples using weight $w_i$. 
To render the final colour, we merge the dense and fine samples and use all in the rendering equation \cite{chen2022tensorf}.

\section{Experiments}
We compare \ours to relevant methods on the commonly used synthetic Blender~\cite{mildenhall2021nerf}, the real-world Tanks and Temples~\cite{knapitsch2017tanks}, and the challenging object-centric Mip-NeRF 360~\cite{barron2022mipnerf360} datasets. To show its effectiveness, we compare \ours to a dense-grid representation and evaluate it with reduced quality of the input point cloud. We start by describing the exact hyperparameters used.

\subsection{Implementation details}\label{sec:impl-details}
\par\noindent\textbf{Generating point cloud \& triangulation.} 
Given a set of posed images, we use the COLMAP reconstruction pipeline \cite{schoenberger2016sfm,schoenberger2016mvs} to get the point cloud used in our tetrahedra field representation.
We then reduce the size of the resulting point cloud such that if the number of points is larger than $10^6$, we subsample $10^6$ points randomly. 
For the Blender dataset experiments, where the number of points is smaller, we add more randomly generated points. In that case, the number of random points is half the number of original points. The reason for adding the points is that with a low number of points, some pixels on edges may not intersect any tetrahedra, potentially producing artefacts on the edges.
Each added point is sampled as follows: we sample a random point $x_0$ from the original point cloud, sample a random normal vector $n$, and a number $\alpha \sim \mathcal{N}(\bar{d},\, \bar{d}^2)$, where $\bar{d}$ is the average spacing of the original point cloud, \ie the average distance between each point and its six closest neighbours. We then add the point $x = x_0 + \alpha n$.

\par\noindent\textbf{Initialization.}
Given the processed point cloud, we apply the Delaunay triangulation \cite{delaunay1934bulletin} to get a set of tetrahedra.
To this end, we use the CGAL library \cite{alliez2022cgal}, which in our experiments runs in the order of milliseconds.
Following \cite{muller2022ingp}, we initialise the features of size $64$ at the vertices of the tetrahedra with small values sampled uniformly in the range $-10^{-4}$ to $10^{-4}$. However, to allow the model to reuse the information contained in the point cloud, we set the first four dimensions of the feature field to the RGBA colour values (rescaled to interval $[0, 1]$) stored at the associated points in the point cloud. The alpha value of all original points is one, whereas all randomly sampled points have an alpha value of zero. For the MLP, we follow the common practice of using the Kaiming uniform initialisation \cite{he2015delving}. The hidden sizes in all MLPs are 128.

\par\noindent\textbf{Training.}
During training, we sample batches of 4,096 random rays from random training images. We use volumetric rendering to predict the colour of each ray. The gradients are computed by backpropagating the MSE loss between the predicted colour and the ground truth colour value. We use the RAdam optimizer \cite{liu2020radam} and decay the learning rate exponentially from $10^{-3}$ to $10^{-4}$ in $300$k steps. The training code is built on top of the Nerfstudio framework \cite{tancik2023nerfstudio}, and the tetrahedra field is implemented in CUDA and uses the OptiX library \cite{parker2010optix}. We train on a single NVIDIA A100 GPU and the training speed ranges from 15k rays per second to 30k rays per second. The speed depends on how well-structured the triangulation is, how many vertices there are, and if there is empty space around the object. The full training with 300k iterations takes between $11$ and $24$ hours, depending on the scene complexity. However, good results are typically obtained much earlier, \eg, in 100k iterations.

\subsection{Results} 
\PAR{Blender dataset~\cite{mildenhall2021nerf} results.}
\begin{table}[t!]
\begin{center}
\small{
    \newcommand{\tfirst}[1]{\cellcolor{tab_red} #1}
    \newcommand{\tsecond}[1]{\cellcolor{orange} #1}
    \newcommand{\tthird}[1]{\cellcolor{yellow} #1}
    \centering
    \begin{tabular}{@{}l|ccc}
& PSNR~$\uparrow$ & SSIM~$\uparrow$ & LPIPS~$\downarrow$ \\ \hline
NeRF\cite{mildenhall2021nerf}& 31.00          & 0.947          & 0.081          \\
NSVF\cite{liu2020nsvf}     & 31.77          & 0.953          & -              \\
mip-NeRF\cite{barron2021mipnerf}                     & \tfirst{34.51} & 0.961          & \tsecond{0.043}\\
instant-NGP\cite{muller2022ingp}                  & \tthird{33.18} & -              & -              \\
Plenoxels\cite{fridovich2022plenoxels}          & 31.71          & 0.958          & \tthird{0.049} \\
Point-NeRF$^{col}$\cite{xu2022pointnerf}          & 31.77          & \tthird{0.973} & 0.062          \\
Point-NeRF$^{mvs}$\cite{xu2022pointnerf}           & \tsecond{33.31}& \tsecond{0.978}& \tthird{0.049} \\
\textbf{\ours}     & 32.53          & \tfirst{0.982} & \tfirst{0.041} \\
\end{tabular}

}%
\end{center}
\caption{\textbf{Results on the Blender dataset}~\cite{mildenhall2021nerf} averaged over all scenes in the dataset. Even though we use the same input point cloud as Point-NeRF$^{col}$, we outperform it greatly. We perform on par with Point-NeRF$^{mvs}$ even though it uses many more points and densifies the point cloud during training. We highlight the \colorbox{tab_red}{best}, \colorbox{orange}{second}, and \colorbox{yellow}{third} values.
\label{tab:blender-small}}
\end{table}
\begin{figure}[t!]
\centering
\small
\begin{tikzpicture}[
 image/.style = {text width=0.24\linewidth, 
                 inner sep=0pt, outer sep=0pt},
label/.style = { minimum height=0.6cm },
node distance = 1pt and 1pt
                        ] 
\path coordinate(last);
    \node [image,below=of last,alias=last] (img00)
    {\includegraphics[width=\linewidth]{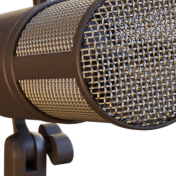}};
\node [image,right=of img00] (img01) 
    {\includegraphics[width=\linewidth]{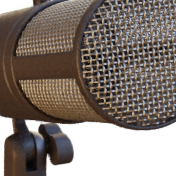}};
\node [image,right=of img01] (img02)
    {\includegraphics[width=\linewidth]{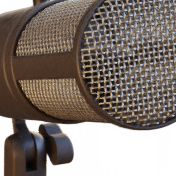}};
\node [image,right=of img02] (img03) 
    {\includegraphics[width=\linewidth]{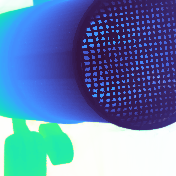}};

    \node [image,below=of last,alias=last] (img10)
    {\includegraphics[width=\linewidth]{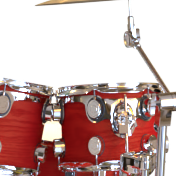}};
\node [image,right=of img10] (img11)
    {\includegraphics[width=\linewidth]{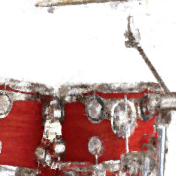}};
\node [image,right=of img11] (img12) 
    {\includegraphics[width=\linewidth]{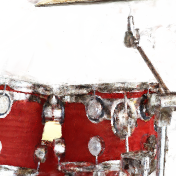}};
\node [image,right=of img12] (img13) 
    {\includegraphics[width=\linewidth]{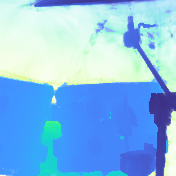}};
    
    \node [image,below=of last,alias=last] (img20)
    {\includegraphics[width=\linewidth]{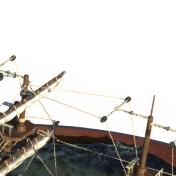}};
\node [image,right=of img20] (img21)
    {\includegraphics[width=\linewidth]{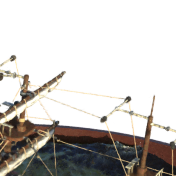}};
\node [image,right=of img21] (img22) 
    {\includegraphics[width=\linewidth]{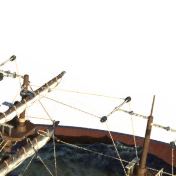}};
\node [image,right=of img22] (img23) 
    {\includegraphics[width=\linewidth]{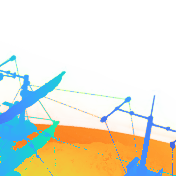}};

    \node [image,below=of last,alias=last] (img30)
    {\includegraphics[width=\linewidth]{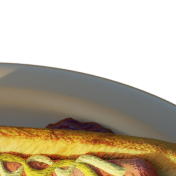}};
\node [image,right=of img30] (img31)
    {\includegraphics[width=\linewidth]{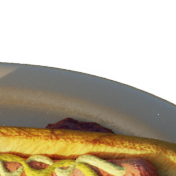}};
\node [image,right=of img31] (img32) 
    {\includegraphics[width=\linewidth]{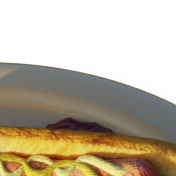}};
\node [image,right=of img32] (img33) 
    {\includegraphics[width=\linewidth]{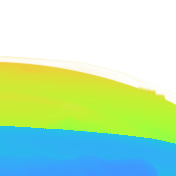}};
    
\node [label,above=-2pt of img00] {ground truth};
\node [label,above=-0pt of img01] {Point-NeRF$^{col}$};
\node [label,above=-2pt of img02] {\textbf{ours}};
\node [label,above=-2pt of img03] {\textbf{ours depth}};
\end{tikzpicture}

\caption{\textbf{Qualitative results on the Blender dataset.}
We compare with Point-NeRF$^{col}$ as we use the same input point cloud.
In the \textbf{top}-row picture, we can see that \ours is able to represent fine details well on the \textit{mic} scene.
On the \textit{drums} scene (\textbf{2$^{\text{nd}}$} row), both methods struggle with shiny materials, but our method performs slightly better.
\textbf{3$^{\text{rd}}$} row: \ours can render thin ropes.
\textbf{Bottom} row: accumulation is nonzero in areas close to surface.
\label{fig:results-blender}
}
\end{figure}
We compare to relevant baselines on %
the standard Blender dataset. %
We used the same split and evaluation procedure as in the original NeRF paper \cite{mildenhall2021nerf} and the same SSIM implementation as Point-NeRF \cite{xu2022pointnerf}.
In order to ensure a fair comparison with Point-NeRF~\cite{xu2022pointnerf} when COLMAP points were used, we use the exact same COLMAP reconstruction as Point-NeRF. We report the PSNR, SSIM, and LPIPS (VGG) \cite{zhang2018lpips} metrics. 
Tab.~\ref{tab:blender-small} shows averaged results, Fig.~\ref{fig:results-blender} shows qualitative results, and
the results for individual scenes are given in \suppmat

When we use the exact same COLMAP points as Point-NeRF (row \textit{Point-NeRF$^{col}$}), we outperform it in all three metrics. We score comparably with \textit{Point-NeRF$^{mvs}$}, even though it starts from a much denser and higher-quality initial point cloud, which it generates from a jointly trained model. Also note that both of these Point-NeRF configurations grow the point cloud during training and, therefore, the complexity of the scene representation grows. For us, the points are fixed, and the number of parameters stays the same. 
We also outperform Plenoxels \cite{fridovich2022plenoxels}, which uses a sparse grid. Note that same as Point-NeRF, Plenoxels also gradually increases the representation complexity by subdividing the grid resolution at predefined training epochs. Even though both Mip-NeRF and instant-NGP outperform our approach in terms of PSNR, our method is slightly better in terms of SSIM, and on par with Mip-NeRF in terms of LPIPS.

\begin{table}[t!]
\begin{center}
\small{
{
\newcommand{\tfirst}[1]{\textbf{#1}}
\newcommand{\tsecond}[1]{#1}
\newcommand{\tthird}[1]{#1}
\setlength\tabcolsep{4pt}
\centering
    \begin{tabular}{@{}l|cc|cc}
    & \multicolumn{2}{c|}{\textit{hotdog}} & \multicolumn{2}{c}{\textit{ship}} \\
& PSNR~$\uparrow$ & SSIM~$\uparrow$ & PSNR~$\uparrow$  & SSIM~$\uparrow$\\ \hline
Point-NeRF$^{static}$        & 29.91 & 0.978  & 19.35 & 0.905 \\
\textbf{\ours}     & 33.31 & 0.989  & 31.13 & 0.994 \\

\end{tabular}
}

}%
\end{center}
\caption{
\textbf{Comparison with Point-NeRF with disabled point cloud growing and prunning} shows that Point-NeRF performs significantly worse in all measured metrics because it struggles to handle the sparse point clouds.
\label{tab:blender-nogrow}
}
\end{table}
To analyze the tetrahedra field representation, we compare it with the point cloud field representation used in Point-NeRF. In Table~\ref{tab:blender-nogrow}, we compare our method to Point-NeRF when we disable point cloud growth and pruning. We show the results on two scenes from the Blender dataset \cite{mildenhall2021nerf} which were selected in the Point-NeRF paper. 
With this setup, we vastly outperform Point-NeRF in all metrics. The reason is that Point-NeRF requires the point cloud to be dense such that all rays have a chance to intersect a neighbourhood of a point. %
Since we use a continuous representation (tetrahedra field) rather than a discrete one, we can achieve good results even for sparser point clouds.

\PAR{Comparison with the dense grid representation.}
\begin{figure}[t!]
\centering
\includegraphics[width=0.475\columnwidth]{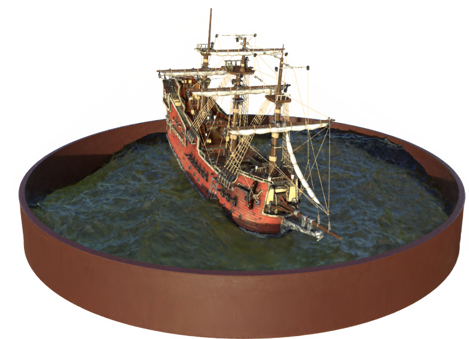}
\includegraphics[width=0.475\columnwidth]{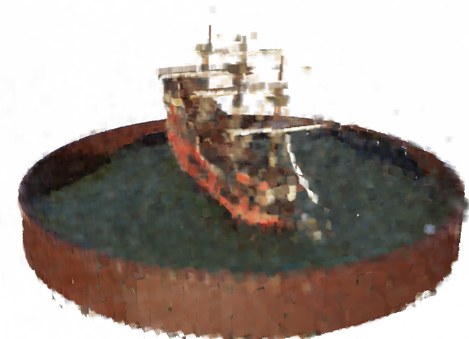}

\caption{\textbf{Comparison with dense grid field representation.}
\ours on the \textbf{left}, dense grid %
on the \textbf{right}. We used a comparable number of parameters for both methods (the dense grid having slightly more parameters). Due to the adaptive nature of tetrahedra fields, \ours produces significantly better rendering as it is able to focus on relevant parts of the scene. 
\label{fig:comparison-dense-grid}}
\end{figure}
In order to show the utility of the adaptive tetrahedra field representation, we compare it to a dense grid representation. Similarly to NSVF \cite{liu2020nsvf}, we split the 3D scene bounding box into equally-sized voxels. When querying the field, we find the voxel to which the query point belongs and perform trilinear interpolation of the eight corners of the voxel. We choose the grid resolution such that the number of grid points is the lowest cube number larger than the number of points of the original point cloud. This ensures a fair comparison as the baseline uses a comparable (but larger) number of features than our approach. All other hyperparameters are kept the same as for \ours. We evaluate both approaches on two scenes from the Blender dataset \cite{mildenhall2021nerf}.
The dense grid representation only scores PSNR $18.81$ and $18.91$ on the \textit{lego} and \textit{ship} scenes, respectively, whereas \ours scores PSNR $33.79$ and $30.69$, respectively.
The results can be seen in Figure~\ref{fig:comparison-dense-grid}.
Note that in this experiment, we only trained for $100,000$ iterations to save computation time. From the numbers and the figure, we can clearly see that the dense grid resolution is not sufficient to reconstruct the scene in enough detail. 
Because \ours uses an adaptive subdivision, which is more detailed around the surface, we are better able to focus on relevant scene parts. With a similar number of trainable parameters, we thus obtain better results.

\PAR{Tanks and Temples~\cite{knapitsch2017tanks} dataset.} %
\begin{table}[t!]
\begin{center}
\small{
    \newcommand{\tfirst}[1]{\cellcolor{tab_red} #1}
    \newcommand{\tsecond}[1]{\cellcolor{orange} #1}
    \newcommand{\tthird}[1]{\cellcolor{yellow} #1}
    \centering
    \begin{tabular}{@{}l|ccc}
    
& PSNR~$\uparrow$ & SSIM~$\uparrow$ & LPIPS~$\downarrow$ \\ \hline
NV~\cite{lombardi2019neural}         & 23.70          & 0.848          & 0.260          \\
NeRF~\cite{mildenhall2021nerf}       & 25.78          & 0.864          & 0.198          \\
NSVF~\cite{liu2020nsvf}              & \tsecond{28.40}& \tthird{0.900} & \tthird{0.153} \\
Point-NeRF~\cite{xu2022pointnerf}$^*$& \tthird{28.35} & \tsecond{0.942}& \tsecond{0.090}\\
\textbf{\ours}             & \tfirst{28.90} & \tfirst{0.957} & \tfirst{0.059} \\
\end{tabular}

}%
\end{center}
\caption{
\textbf{Result on the Tanks and Temples dataset \cite{knapitsch2017tanks}} as processed by NSVF \cite{liu2020nsvf} averaged over all scenes. We show the PSNR, SSIM, and LPIPS (Alex) \cite{zhang2018lpips} metrics.
We highlight the \colorbox{tab_red}{best}, \colorbox{orange}{second}, and \colorbox{yellow}{third} values.
\textit{$^*$Note, that Point-NeRF \cite{xu2022pointnerf} results differ from the paper as they were recomputed with the resolution used in other methods.}
\label{tab:tt-small}}
\end{table}
\begin{figure}[t!]
\centering
\small
\begin{tikzpicture}[
 image/.style = {text width=0.32\linewidth, 
                 inner sep=0pt, outer sep=0pt},
label/.style = { minimum height=0.4cm },
node distance = 1pt and 1pt
                        ] 
\path coordinate(last);
    \node [image,below=of last,alias=last] (img00)
    {\includegraphics[width=\linewidth]{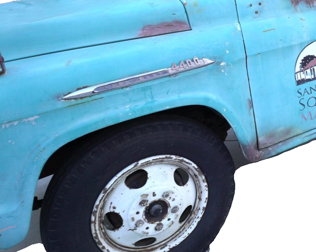}};
\node [image,right=of img00] (img01)
    {\includegraphics[width=\linewidth]{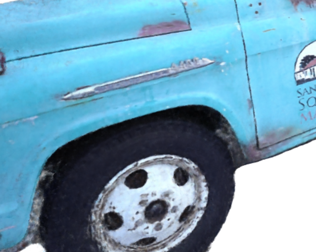}};
\node [image,right=of img01] (img02) 
    {\includegraphics[width=\linewidth]{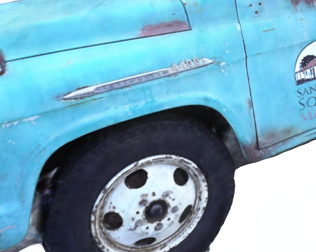}};
    \draw[draw=red,line width=0.5mm] ($ (img00) + (-0.8,-0.94) $) rectangle ++(0.6,0.6);
    \draw[draw=red,line width=0.5mm] ($ (img01) + (-0.8,-0.94) $) rectangle ++(0.6,0.6);
    \draw[draw=red,line width=0.5mm] ($ (img02) + (-0.8,-0.94) $) rectangle ++(0.6,0.6);

    \node [image,below=of last,alias=last] (img10)
    {\includegraphics[width=\linewidth]{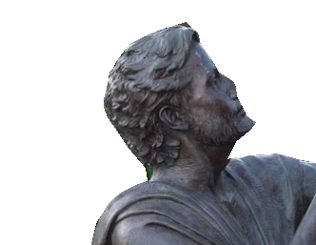}};
\node [image,right=of img10] (img11)
    {\includegraphics[width=\linewidth]{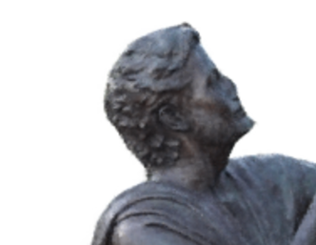}};
\node [image,right=of img11] (img12) 
    {\includegraphics[width=\linewidth]{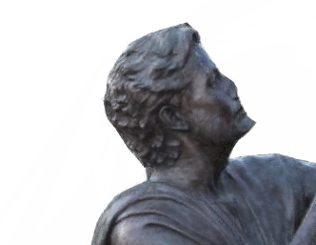}};
    \draw[draw=red,line width=0.5mm] ($ (img10) + (-0.3,-0.0) $) rectangle ++(0.6,0.6);
    \draw[draw=red,line width=0.5mm] ($ (img11) + (-0.3,-0.0) $) rectangle ++(0.6,0.6);
    \draw[draw=red,line width=0.5mm] ($ (img12) + (-0.3,-0.0) $) rectangle ++(0.6,0.6);
    
    \node [image,below=of last,alias=last] (img20)
    {\includegraphics[width=\linewidth]{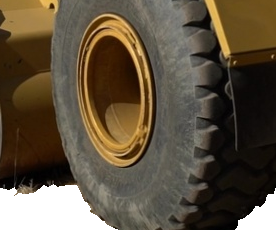}};
\node [image,right=of img20] (img21)
    {\includegraphics[width=\linewidth]{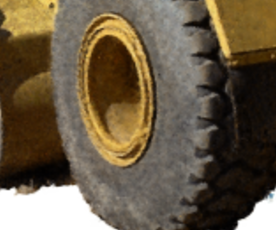}};
\node [image,right=of img21] (img22) 
    {\includegraphics[width=\linewidth]{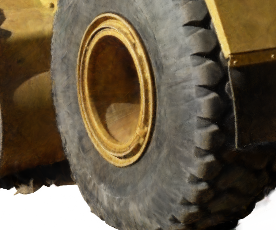}};
    \draw[draw=red,line width=0.5mm] ($ (img20) + (-0.8,-0.82) $) rectangle ++(0.6,0.6);
    \draw[draw=red,line width=0.5mm] ($ (img21) + (-0.8,-0.82) $) rectangle ++(0.6,0.6);
    \draw[draw=red,line width=0.5mm] ($ (img22) + (-0.8,-0.82) $) rectangle ++(0.6,0.6);
    
    \node [image,below=of last,alias=last] (img30)
    {\includegraphics[width=\linewidth]{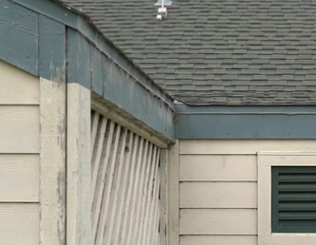}};
\node [image,right=of img30] (img31)
    {\includegraphics[width=\linewidth]{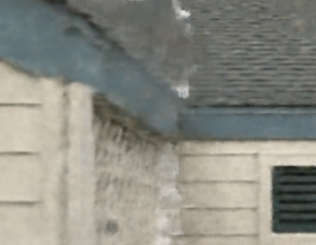}};
\node [image,right=of img31] (img32) 
    {\includegraphics[width=\linewidth]{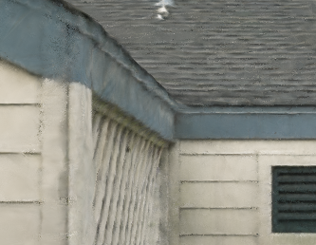}};
    \draw[draw=red,line width=0.5mm] ($ (img30) + (-0.8,-0.78) $) rectangle ++(0.6,0.6);
    \draw[draw=red,line width=0.5mm] ($ (img31) + (-0.8,-0.78) $) rectangle ++(0.6,0.6);
    \draw[draw=red,line width=0.5mm] ($ (img32) + (-0.8,-0.78) $) rectangle ++(0.6,0.6);
    
\node [label,above=-1pt of img00] {ground truth};
\node [label,above=-1pt of img01] {Point-NeRF};
\node [label,above=-1pt of img02] {Tetra-NeRF};
\end{tikzpicture}

\caption{\textbf{Results on Tanks and Temples dataset.}
In the \textbf{top} row, we can see that we are able to represent the rim of the wheel better than Point-NeRF.
Similarly, we can represent the tyre better in the \textbf{third row}. Finally, in the \textbf{bottom}, Point-NeRF fails on the wall and the roof, whereas \ours can render these parts well.
\label{fig:results-tt}
}
\end{figure}
To be able to compare with Point-NeRF \cite{xu2022pointnerf} on real-world data, we have evaluated \ours on the Tanks and Temples~\cite{knapitsch2017tanks} dataset. We use the same setup as in NSVF \cite{liu2020nsvf}, where the object is masked.
We report the usual metrics: PSNR, SSIM, and LPIPS (Alex).\footnote{
We always choose the type of LPIPS (Alex \cite{krizhevsky2012imagenet}/VGG \cite{simonyan2014very}) such that we can compare with more methods as some only evaluate using one type.} 
We used the dense COLMAP reconstruction to get the point cloud used in the tetrahedra field. Quantitative results are shown in Table~\ref{tab:tt-small}, qualitative results are presented in Figure~\ref{fig:results-tt}, and the per-scene results are given in the \suppmat
Note that the results originally reported in the Point-NeRF paper \cite{xu2022pointnerf} were evaluated with a resolution different from the compared methods. Therefore, to ensure a fair comparison, we have recomputed the metrics in the publicly available Point-NeRF's predictions with the full resolution of $1920 \times 1080$ as used in NSVF \cite{liu2020nsvf}. In the public dataset, one of the scenes had corrupted camera parameters and we had to reconstruct the poses again for the Ignatius scene. With the reconstructed poses, our method performs slightly worse on that scene compared to others. 

The results show %
that our method outperforms the baselines in all compared metrics.
This indicates that even though Point-NeRF relies on the ability to grow the point cloud density during training, this is not needed when using a continuous instead of a discrete representation. %

\begin{table*}[hbt!]
\begin{center}
\footnotesize{
    \newcommand{\tfirst}[1]{\cellcolor{tab_red} #1}
    \newcommand{\tsecond}[1]{\cellcolor{orange} #1}
    \newcommand{\tthird}[1]{\cellcolor{yellow} #1}
    \centering
\setlength\tabcolsep{5pt}
    \small
    \begin{tabular}{@{}l|ccc|ccc|ccc}
    & \multicolumn{3}{c|}{\textit{Outdoor}} & \multicolumn{3}{c|}{\textit{Indoor}} & \multicolumn{3}{c}{\textit{Mean}} \\
& PSNR~$\uparrow$ & SSIM~$\uparrow$ & LPIPS~$\downarrow$& PSNR~$\uparrow$ & SSIM~$\uparrow$ & LPIPS~$\downarrow$& PSNR~$\uparrow$ & SSIM~$\uparrow$ & LPIPS~$\downarrow$ \\ \hline 
NeRF~\cite{mildenhall2021nerf, jaxnerf2020github}   & 21.46          & 0.458          & 0.515          & 26.84          & 0.790          & 0.370          & 23.85          & 0.605          & 0.451          \\
mip-NeRF~\cite{barron2021mipnerf}                   & 21.69          & 0.471          & 0.505          & 26.98          & 0.798          & 0.361          & 24.04          & 0.616          & 0.441          \\
NeRF++~\cite{zhang2020nerf++}                       & 22.76          & 0.548          & 0.427          & 28.05          & 0.836          & 0.309          & 25.11          & 0.676          & 0.375          \\
Deep Blending~\cite{hedman2018deep}                 & 21.54          & 0.524          & 0.364          & 26.39          & 0.844          & 0.261          & 23.70          & 0.666          & 0.318          \\
Point-Based Neural Rendering~\cite{kopanas2021point}& 21.66          & \tthird{0.612} & 0.302          & 26.28          & \tthird{0.887} & 0.191          & 23.71          & \tthird{0.734} & \tthird{0.253} \\
Stable View Synthesis~\cite{riegler2021stable}      & \tthird{23.01} & \tsecond{0.662}& \tfirst{0.253} & \tthird{28.22} & \tsecond{0.907}& \tsecond{0.160}& \tthird{25.33} & \tsecond{0.771}& \tfirst{0.211} \\
mip-NeRF 360~\cite{barron2022mipnerf360}            & \tfirst{23.72} & \tfirst{0.687} & \tsecond{0.282}& \tsecond{29.43}& \tfirst{0.911} & \tthird{0.181} & \tsecond{26.26}& \tfirst{0.786} & \tsecond{0.237}\\
\textbf{\ours}                                      & \tsecond{23.17}& 0.586          & \tthird{0.298} & \tfirst{30.21} & 0.881          & \tfirst{0.103} & \tfirst{26.30} & 0.717          & \tfirst{0.211} \\
    \end{tabular}

}
\end{center}
\caption{\textbf{Mip-NeRF 360 dataset results.}
We show the PSNR, SSIM, and LPIPS (Alex) \cite{zhang2018lpips} on two categories of Mip-NeRF 360 \cite{barron2022mipnerf360} scenes: \textit{outdoor}, and \textit{indoor}.
On the \textit{outdoor} scenes, we outperform the Stable View Synthesis \cite{riegler2021stable}
 and are comparable to Mip-NeRF 360 \cite{barron2022mipnerf360}, even though our method does not implement the improvements suggested in NeRF++ \cite{zhang2020nerf++} and Mip-NeRF 360 \cite{barron2022mipnerf360}, and is more comparable to vanilla NeRF \cite{mildenhall2021nerf}.
 We highlight the \colorbox{tab_red}{best}, \colorbox{orange}{second}, and \colorbox{yellow}{third} values.
 \label{tab:mipnerf360-small}}
\end{table*}

\PAR{Sparse and dense point cloud comparison.}
\begin{figure}[t!]
\centering
\subfloat[\centering sparse point cloud]{{
    \includegraphics[width=0.4\columnwidth]{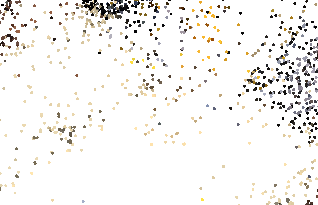}
}}
\subfloat[\centering ground-truth image]{{
    \includegraphics[width=0.4\columnwidth]{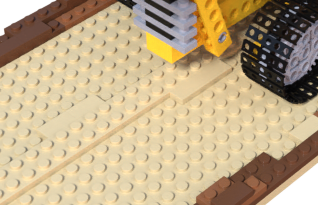}
}}
\\
\subfloat[\centering sparse model]{{
    \includegraphics[width=0.40\columnwidth]{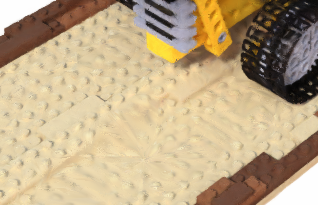}
}}        
\subfloat[\centering dense model]{{
    \includegraphics[width=0.40\columnwidth]{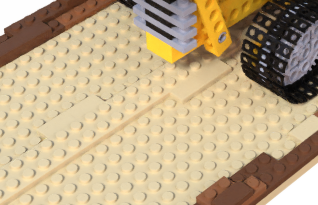}
}}

\caption{\textbf{Sparse \vs dense model comparison detail.} %
Even though the area in the bottom %
contains few points, we are still able to reconstruct at least low-frequency data.
\label{fig:sparse-model-detail}
}
\end{figure}
\begin{table}[t!]
\begin{center}
\small{
\centering
    \newcommand{\tfirst}[1]{\cellcolor{tab_red} #1}
    \newcommand{\tsecond}[1]{\cellcolor{orange} #1}
    \newcommand{\tthird}[1]{\cellcolor{yellow} #1}
\setlength\tabcolsep{2pt}
    \centering
    \begin{tabular}{@{}l|cc|cc}
& \multicolumn{2}{c|}{\textit{sparse}} & \multicolumn{2}{c}{\textit{dense}} \\
& PSNR/SSIM & \#points & PSNR/SSIM  & \#points\\ \hline
Blender/lego         & 29.77/0.959 & 25,784   & 33.79/0.985  &  302,781 \\
Blender/ship         & 26.90/0.899 & 7,152    & 30.69/0.942  &  321,861 \\
\hline
360/bonsai           & 28.12/0.902 & 413,226  & 28.34/0.902 & 1,000,000 \\
360/garden           & 24.79/0.806 & 227,532  & 25.41/0.838 & 1,000,000   \\
\end{tabular}

}%
\end{center}
\vspace{0.3\baselineskip}
\caption{\textbf{Dense and sparse point cloud comparison.} We present results on two scenes from the Blender \cite{mildenhall2021nerf} and Mip-NeRF 360 \cite{barron2022mipnerf360} datasets. We also show the number of vertices of the tetrahedra field -- including the randomly sampled points.
\label{tab:dense-sparse-comparison}
}
\end{table}
Previous experiments have used dense COLMAP reconstructions. 
However, dense point clouds may not be required to achieve good reconstructions. In this section, we compare models trained on dense reconstructions and sparse reconstructions. Table~\ref{tab:dense-sparse-comparison} compares PSNR, and SSIM metrics on two synthetic scenes from the Blender dataset \cite{mildenhall2021nerf} and two real-world scenes from the Mip-NeRF 360 dataset \cite{barron2022mipnerf360}.
We also show the number of vertices of the tetrahedra, which is the size of the input point cloud, including the random points added at the beginning.
Note that in this experiment, we only trained for $100,000$ iterations to save computation time.

From the results, we can see that for the Blender dataset~\cite{mildenhall2021nerf} case, the dense model is much better. This is to be expected since the number of points obtained from the sparse reconstruction on the Blender dataset is very small and our MLP does not have enough capacity to represent fine details. However, even in regions with zero point coverage, we are still able to provide at least some low-frequency data (\cf Fig.~\ref{fig:sparse-model-detail}). 
On real-world scenes, the sparse point cloud model is almost on par with the dense one. The sparse reconstructions on the real-world dataset provide many more points compared to the synthetic dataset case, and the coverage is sufficient.
From these results, we conclude that on real-world data the sparse model can be sufficient to achieve good performance.

\PAR{Mip-NeRF 360\cite{barron2022mipnerf360} dataset.}
\begin{figure}[t!]
\centering
\small
\begin{tikzpicture}[
 image/.style = {text width=0.32\linewidth, 
                 inner sep=0pt, outer sep=0pt},
label/.style = { minimum height=0.4cm },
node distance = 1pt and 1pt
                        ] 
\path coordinate(last);
    \node [image,below=of last,alias=last] (img00)
    {\includegraphics[width=\linewidth]{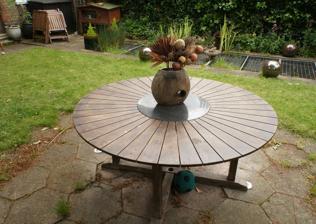}};
\node [image,right=of img00] (img01)
    {\includegraphics[width=\linewidth]{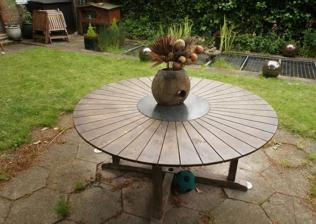}};
\node [image,right=of img01] (img02)
    {\includegraphics[width=\linewidth]{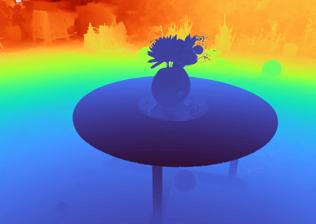}};

    \node [image,below=of last,alias=last] (img10)
    {\includegraphics[width=\linewidth]{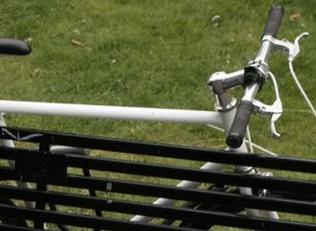}};
\node [image,right=of img10] (img11)
    {\includegraphics[width=\linewidth]{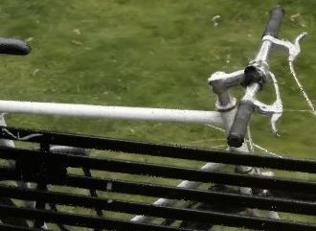}};
\node [image,right=of img11] (img12) 
    {\includegraphics[width=\linewidth]{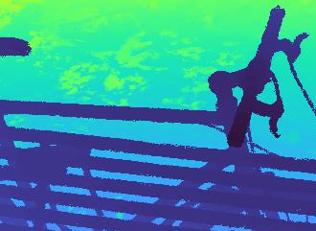}};
    
    \node [image,below=of last,alias=last] (img20)
    {\includegraphics[width=\linewidth]{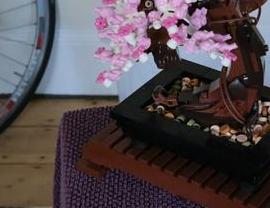}};
\node [image,right=of img20] (img21)
    {\includegraphics[width=\linewidth]{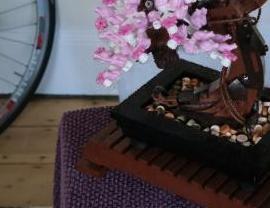}};
\node [image,right=of img21] (img22) 
    {\includegraphics[width=\linewidth]{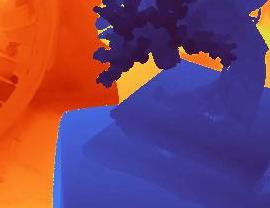}};

    \node [image,below=of last,alias=last] (img30)
    {\includegraphics[width=\linewidth]{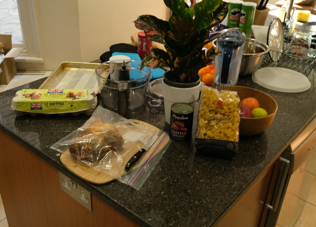}};
\node [image,right=of img30] (img31)
    {\includegraphics[width=\linewidth]{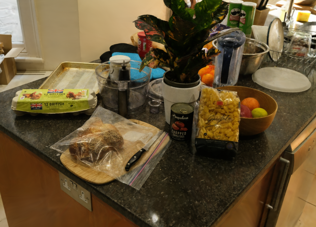}};
\node [image,right=of img31] (img32)
    {\includegraphics[width=\linewidth]{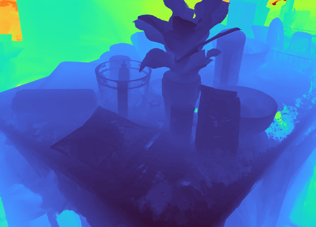}};

\node [label,above=-1pt of img00] {ground truth};
\node [label,above=-1pt of img01] {prediction};
\node [label,above=-1pt of img02] {depth};
\end{tikzpicture}

\caption{\textbf{Mip-NeRF 360 results.} \textbf{1$^{\text{st}}$ row}: we show the results on the \textit{garden} scene. \textbf{2$^{\text{nd}}$}: we can see that \ours has some problems rendering grass. \textbf{3$^{\text{rd}}$}: \ours is able to represent the delicate texture of the tablecloth well. \textbf{4$^{\text{th}}$}: \ours is able to recover fine geometry. %
\label{fig:results-360}
}
\end{figure}
We have further evaluated our method on the Mip-NeRF 360~\cite{barron2022mipnerf360} dataset. 
In order to ensure a fair comparison with Mip-NeRF 360~\cite{barron2022mipnerf360}, we trained and evaluated on four times downsized images for the outdoor scenes and two times downsized images for the indoor scenes.
The quantitative results are presented in Table~\ref{tab:mipnerf360-small}, and  the qualitative results are shown in Figure~\ref{fig:results-360}.

From the results, we can see that \ours is able to outperform both the vanilla NeRF \cite{mildenhall2021nerf} and Mip-NeRF \cite{barron2021mipnerf}. We also outperform Stable View Synthesis \cite{riegler2021stable} in terms of PSNR and score comparably in terms of LPIPS. This is possible because outdoor scenes contain a lot of high-resolution geometries, such as leaves and Stable View Synthesis is not able to overcome noise in the approximate geometry. \ours does not suffer from these problems as we aggregate features in 3D along rays rather than on the surface.
Mip-NeRF 360 scores comparably to \ours in terms of PSNR and LPIPS, but it outperforms \ours in terms of SSIM. However, Mip-NeRF 360 implements some tricks designed to boost its performance. On the other hand, our approach is based on vanilla NeRF, and we
believe that the performance can be boosted similarly. %

\PAR{Varying the number of input points}\label{sec:ablation-num-points}
\definecolor{plt-blue}{HTML}{1f77b4}
\definecolor{plt-orange}{HTML}{ff7f0e}
\definecolor{plt-green}{HTML}{2ca02c}
\definecolor{plt-red}{HTML}{d62728}
We conducted a study on the effect of the input point cloud size on the reconstruction quality. We consider both the sparse and the dense point clouds and analyse the performance as we sub-sample the points or add more points randomly. We evaluate on the
\textcolor{plt-blue}{\textit{family}} and the \textcolor{plt-orange}{\textit{truck}} scenes from the Tanks and Temples~\cite{knapitsch2017tanks} (tt) dataset, and the \textcolor{plt-red}{\textit{room}} and the \textcolor{plt-green}{\textit{garden}} scenes from the
Mip-NeRF 360 \cite{barron2022mipnerf360} dataset. The sizes of sparse point clouds were roughly 16K, 30K, 113K, and 139K respectively and the sizes of dense point clouds were larger than 5M for all scenes.
When the size of the original point cloud was larger than the desired size, we selected points uniformly at random, and when smaller, we added new random points as described Section~\ref{sec:impl-details}.
We include detailed results in the \suppmat In order to save computational resources, we only train the method for $100$k iterations. The results are visualised in Figure~\ref{fig:ablation-num-input-points}.

As expected, the performance improves with the number of points used, as it leads to a finer subdivision of the scene around the surface. We can also observe (\eg, from \textit{tt/family}) that the sparse reconstruction leads to a better reconstruction quality at the same point cloud size. The likely explanation is that sampling uniformly at random will be biased towards regions with high density and some regions may be missed or covered extremely sparsely.

\PAR{Speed of convergence}
\definecolor{plt-blue}{HTML}{1f77b4}
\definecolor{plt-orange}{HTML}{ff7f0e}
\definecolor{plt-green}{HTML}{2ca02c}
\definecolor{plt-red}{HTML}{d62728}
\begin{figure*}[ht!]
\centering
\subfloat[\centering Different point cloud sizes (PSNR)\label{fig:ablation-num-input-points}]{{
    \includegraphics[width=0.32\textwidth]{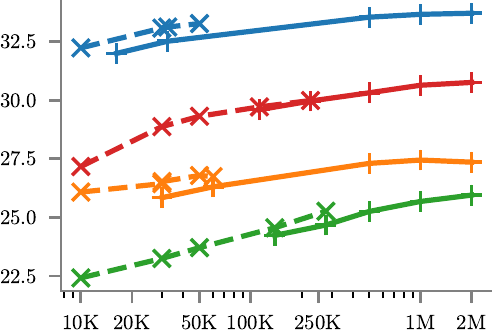}
}}        
\subfloat[\centering Training performance on blender/ship (PSNR)\label{fig:convergence-speed-1}]{{
\includegraphics[width=0.32\textwidth]{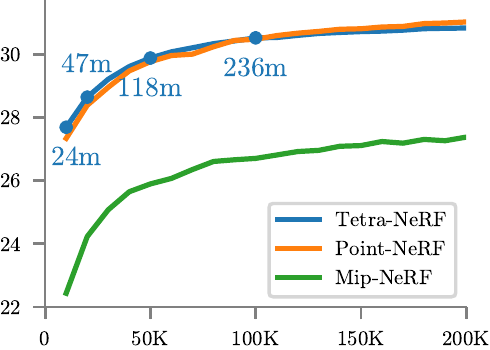}
}}
\subfloat[\centering Training performance (PSNR)\label{fig:convergence-speed-2}]{{
\includegraphics[width=0.32\textwidth]{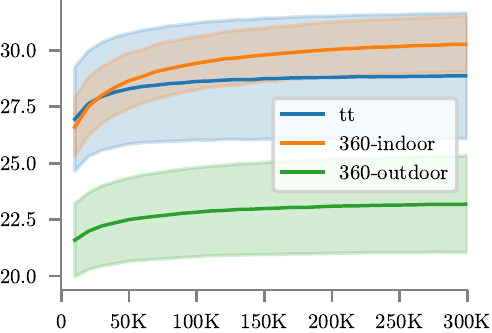}
}}
\vspace{0.4\baselineskip} %
\caption{%
\textbf{a)} shows the PSNR with different sizes of input point clouds. The solid and dashed lines represent the dense and coarse COLMAP reconstructions. The following scenes were evaluated (best to worst): \textcolor{plt-blue}{\textit{tt/family}}, \textcolor{plt-red}{\textit{360/room}}, \textcolor{plt-orange}{\textit{tt/truck}}, and \textcolor{plt-green}{\textit{360/garden}} from Tanks and Temples~\cite{knapitsch2017tanks} (tt) and Mip-NeRF 360~
\cite{barron2022mipnerf360} (360-indoor/360-outdoor) datasets. As expected, the quality increases with the number of points.
\textbf{b)} visualizes the training speed (PSNR at different training iterations) on the \textit{blender/ship} scene \cite{mildenhall2021nerf}. It shows Tetra-NeRF training times for various timesteps and compares the rate of convergence with Point-NeRF~\cite{xu2022pointnerf} and Mip-NeRF~\cite{barron2021mipnerf}.
Similarly, \textbf{c)} shows the rate of convergence aggregated over Tanks and Temples~\cite{knapitsch2017tanks} (tt) and Mip-NeRF 360~
\cite{barron2022mipnerf360} (360-indoor/360-outdoor) datasets. It shows the average PSNR, and the highlighted area represents the standard deviation.
\label{fig:convergence-speed}
}
\end{figure*}
Figures~\ref{fig:convergence-speed-1},~\ref{fig:convergence-speed-2} show the rate of convergence on the \textit{ship} scene from the Blender dataset and aggregated results on the Tanks and Temples \cite{knapitsch2017tanks}, and Mip-NeRF 360 (indoor/outdoor) \cite{barron2022mipnerf360} datasets.
Similar to PointNeRF, good results can be achieved quite early in the training process.
\Eg, on the \textit{blender/ship} scene, in 24 minutes, our method achieves performance similar to Mip-NeRF~\cite{barron2021mipnerf} after several hours of training.
The efficiency of the implementation is a large factor in the speed of different methods.
\Eg, in Instant-NGP \cite{muller2022ingp}, the authors invested a significant effort in optimizing GPU memory accesses, and the result is highly optimised efficient code (a main contribution of~\cite{muller2022ingp}).
There are several places where our implementation can easily be made more efficient. 
\Eg, changing the memory layout (row-major \vs column-major)
of the tetrahedra field leads to a 10\% speedup.

\section{Limitations}
\begin{figure}
\centering
\subfloat[\centering \textbf{Failure case 1}: low point cloud density on the ground\label{fig:failure-cases-a}]{{
    \includegraphics[width=0.49\columnwidth]{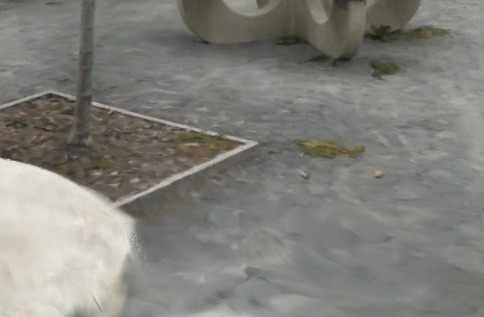}
}}
\subfloat[\centering \textbf{Failure case 2}: too many intersected tetrahedra\label{fig:failure-cases-b}]{{
    \includegraphics[width=0.49\columnwidth]{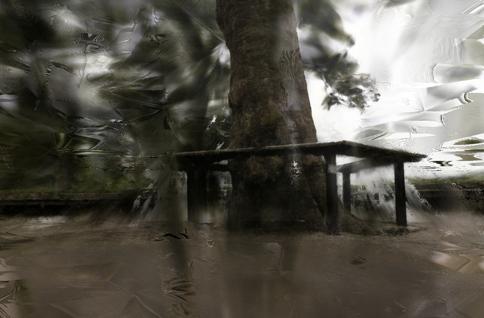}
}}

\caption{\textbf{Failure cases.} 
\textbf{Left}: \textit{barn} scene from~\cite{knapitsch2017tanks}. %
\textbf{Right}: \textit{tree hill} scene from~%
\cite{barron2022mipnerf360}.
\label{fig:failure-cases}
}
\end{figure}
One drawback of our approach is that the quality of different regions of the rendered scene depends on the density of the point cloud in these regions. If the original point cloud was constructed using 2D feature matching, there can be regions with a very low number of 3D points. We show such an example in Fig.~\ref{fig:failure-cases-a}. There are few 3D points on the ground, resulting in blurry renderings for that region. %

Another issue is that the current implementation has a limit on the number of intersected tetrahedra per ray. For large or badly structured scenes, where rays intersect many tetrahedra, this can limit the reconstruction quality. Fig.~\ref{fig:failure-cases-b} shows such a case on a Mip-NeRF 360 scene~\cite{barron2022mipnerf360}. This problem can be addressed by increasing the limit on the number of visited tetrahedra at the cost of higher memory requirements and run-time. 
Alternatively, coarse-to-fine schemes that start with a coarser tetrahedralisation and prune the space could potentially handle this issue.

\section{Conclusion}
This paper proposes a novel radiance field representation that, compared to standard voxel-based representations, is easily able to adapt to 3D geometry priors given in the form of a (sparse) point cloud. 
Our approach elegantly combines concepts from geometry processing (Delaunay triangulation) and triangle-based rendering (ray-triangle intersections) with modern neural rendering approaches. 
The representation has a naturally higher resolution in the space near surfaces, and the input point cloud provides a straightforward way to initialise the radiance field. 
Compared to Point-NeRF, a state-of-the-art point cloud-based radiance field representation which uses the same input as our approach,
\ours
shows clearly better results. 
Our method %
performs comparably to state-of-the-art MLP-based methods. 
The 
results demonstrate
that \ours is an interesting alternative to existing radiance field representations that is worth further investigation.  
Interesting research directions include adaptive refinement and pruning of the tetrahedralisation and exploiting the fact that the surface of the scene is likely close to some of the triangles in the scene.

{
\PAR{Acknowledgements}
This work was supported by the Czech Science Foundation (GA\v{C}R) EXPRO (grant no. 23-07973X),
the Grant Agency of the Czech Technical University in Prague (grant no. SGS22/112/OHK3/2T/13),
and by the Ministry of Education, Youth and Sports of the Czech Republic through the e-INFRA CZ (ID:90254).
}

{
\onecolumn
\appendix
\clearpage
\makeatletter
   \vskip .375in
   \begin{center}
      {\Large \bf Supplementary Material \par}
      \vspace*{24pt}
      {
      \large
      \lineskip .5em
      \begin{tabular}[t]{c}
         \ificcvfinal\@author\else Anonymous ICCV submission\\
         \vspace*{1pt}\\
Paper ID \iccvPaperID \fi
      \end{tabular}
      \par
      }
      \vskip .5em
      \vspace*{12pt}
   \end{center}
\makeatother

First, in Section~\ref{sec:video}, we describe the \textbf{attached video}\footnote{Video can be found at: \url{https://jkulhanek.com/tetra-nerf/video.html}} where we illustrate the tetrahedra field and the optimisation process. Next, we extend Sections~4.2, 4.3, and 4.4 from the main paper by giving detailed results on the Blender \cite{mildenhall2021nerf}, Tanks and Temples \cite{knapitsch2017tanks}, and Mip-NeRF 360 \cite{barron2022mipnerf360} datasets in Sections \ref{sec:blender}, \ref{sec:tt}, and \ref{sec:mipnerf360}. Finally, we extend Section~4.5 from the main paper and evaluate how the performance changes when varying the size of the input point cloud in Section~\ref{sec:ablation-num-points}.

\section{Attached video}\label{sec:video}
To present the idea of the paper visually, we include a video illustrating the tetrahedra field representation and showing the results of the optimisation at different points in the early stages of the training. In the video, we used the \textit{garden} scene from the Mip-NeRF 360 dataset \cite{barron2022mipnerf360}. We show the optimisation using both the sparse and the dense input point clouds. The video starts by showing the initial point cloud and the tetrahedra obtained by the Delaunay triangulation. It then shows how the scene is optimised from the first iteration. Finally, it presents the resulting video generated by the fully trained model.

\section{Blender results}\label{sec:blender}
\begin{table*}[b!]
\small
    \newcommand{\tfirst}[1]{\cellcolor{tab_red} #1}
    \newcommand{\tsecond}[1]{\cellcolor{orange} #1}
    \newcommand{\tthird}[1]{\cellcolor{yellow} #1}
      \setlength\tabcolsep{4pt}
      \captionsetup{aboveskip=5pt}
      \centering
      \begin{tabular}{@{}l|cccccccc|c}
          \multicolumn{1}{c}{\,} & \multicolumn{9}{c}{PSNR$\uparrow$}\\
                       & \textit{chair}          & \textit{drums}        & \textit{ficus} & \textit{hotdog} &   \textit{lego}               & \textit{materials}   & \textit{mic}          & \textit{ship}      & \textit{mean}  \\ \hline
NPBG~\cite{aliev2020neural}   & 26.47          & 21.53          & 24.60          & 29.01          & 24.84          & 21.58          & 26.62          & 21.83          & 24.56          \\
NeRF~\cite{mildenhall2021nerf}& 33.00          & 25.01          & 30.13          & 36.18          & 32.54          & 29.62          & 32.91          & 28.65          & 31.01          \\
NSVF~\cite{liu2020nsvf}     & 33.19          & 25.18          & 31.23          & 37.14          & 32.54          & \tfirst{32.68} & 34.27          & 27.93          & 31.77          \\
Mip-NeRF~\cite{barron2021mipnerf}        & \tfirst{37.14} & \tfirst{27.02} & 33.19          & \tfirst{39.31} & \tsecond{35.74}& \tsecond{32.56}& \tfirst{38.04} & \tfirst{33.08} & \tfirst{34.51} \\
Instant-NGP~\cite{muller2022ingp}       & 35.00          & \tthird{26.02} & \tsecond{33.51}& \tsecond{37.40}& \tfirst{36.39} & \tthird{29.78} & \tsecond{36.22}& \tthird{31.10} & \tthird{33.18} \\
Plenoxels~\cite{fridovich2022plenoxels}                   & 33.98          & 25.35          & 31.83          & 36.43          & 34.10          & 29.14          & 33.26          & 29.62          & 31.71          \\
Point-NeRF$^{col}$~\cite{xu2022pointnerf}           & \tthird{35.09} & 25.01          & 33.24          & 35.49          & 32.65          & 26.97          & 35.54          & 30.18          & 31.77          \\
Point-NeRF$^{mvs}$~\cite{xu2022pointnerf}           & \tsecond{35.40}& \tsecond{26.06}& \tfirst{36.13} & \tthird{37.30} & \tthird{35.04} & 29.61          & \tthird{35.95} & 30.97          & \tsecond{33.31}\\
\textbf{\ours}     & 35.05          & 25.01          & \tthird{33.31} & 36.16          & 34.75          & 29.30          & 35.49          & \tsecond{31.13}& 32.53          \\
\hline
\multicolumn{10}{c}{}\\
           \multicolumn{1}{c}{\,} & \multicolumn{9}{c}{SSIM$\uparrow$}\\
                       & \textit{chair}          & \textit{drums}        & \textit{ficus} & \textit{hotdog} &   \textit{lego}               & \textit{materials}   & \textit{mic}          & \textit{ship}      & \textit{mean}  \\ \hline
NPBG~\cite{aliev2020neural}                  & 0.939          & 0.904          & 0.940          & 0.964          & 0.923          & 0.887          & 0.959          & 0.866          & 0.923          \\
NeRF~\cite{mildenhall2021nerf}                  & 0.967          & 0.925          & 0.964          & 0.974          & 0.961          & 0.949          & 0.980          & 0.856          & 0.947          \\
NSVF~\cite{liu2020nsvf}                  & 0.968          & 0.931          & 0.973          & 0.980          & 0.960          & \tsecond{0.973}& 0.987          & 0.854          & 0.953          \\
Mip-NeRF~\cite{barron2021mipnerf}              & \tthird{0.981} & 0.932          & \tthird{0.980} & 0.982          & 0.978          & 0.959          & \tthird{0.991} & 0.882          & 0.961          \\
Plenoxels~\cite{fridovich2022plenoxels}             & 0.977          & 0.933          & 0.890          & 0.985          & 0.976          & \tfirst{0.975} & 0.980          & \tsecond{0.949}& 0.958          \\
Point-NeRF$^{col}$~\cite{xu2022pointnerf}     & \tsecond{0.990}& \tthird{0.944} & \tsecond{0.989}& \tthird{0.986} & \tthird{0.983} & 0.955          & \tsecond{0.993}& 0.941          & \tthird{0.973} \\
Point-NeRF$^{mvs}$~\cite{xu2022pointnerf}     & \tfirst{0.991} & \tfirst{0.954} & \tfirst{0.993} & \tfirst{0.991} & \tfirst{0.988} & \tthird{0.971} & \tfirst{0.994} & \tthird{0.942} & \tsecond{0.978}\\
\textbf{\ours}         & \tsecond{0.990}& \tsecond{0.947}& \tsecond{0.989}& \tsecond{0.989}& \tsecond{0.987}& 0.968          & \tsecond{0.993}& \tfirst{0.994} & \tfirst{0.982} \\
\hline

\multicolumn{10}{c}{}\\
\multicolumn{1}{c}{\,} & \multicolumn{9}{c}{LPIPS$\downarrow$}\\
& \textit{chair}          & \textit{drums}        & \textit{ficus} & \textit{hotdog} &   \textit{lego}               & \textit{materials}   & \textit{mic}          & \textit{ship}      & \textit{mean}  \\ \hline
NPBG~\cite{aliev2020neural}                   & 0.085          & 0.112          & 0.078          & 0.075          & 0.119          & 0.134          & 0.060          & 0.210          & 0.109          \\
NeRF~\cite{mildenhall2021nerf}                   & 0.046          & 0.091          & 0.044          & 0.121          & 0.050          & 0.063          & 0.028          & 0.206          & 0.081          \\
Mip-NeRF~\cite{barron2021mipnerf}                 & \tsecond{0.021}& \tfirst{0.065} & \tfirst{0.020} & \tfirst{0.027} & \tfirst{0.021} & \tfirst{0.040} & \tfirst{0.009} & 0.138          & \tsecond{0.043}\\
Plenoxels~\cite{fridovich2022plenoxels}              & 0.031          & \tsecond{0.067}& 0.026          & \tsecond{0.037}& 0.028          & \tthird{0.057} & 0.015          & \tthird{0.134} & \tthird{0.049} \\
Point-NeRF$^{col}$~\cite{xu2022pointnerf}     & 0.026          & 0.099          & 0.028          & \tthird{0.061} & 0.031          & 0.100          & 0.019          & \tthird{0.134} & 0.062          \\
Point-NeRF$^{mvs}$~\cite{xu2022pointnerf}      & \tthird{0.023} & 0.078          & \tsecond{0.022}& \tsecond{0.037}& \tthird{0.024} & 0.072          & \tthird{0.014} & \tsecond{0.124}& \tthird{0.049} \\
\textbf{\ours}& \tfirst{0.016} & \tthird{0.073} & \tthird{0.023} & \tfirst{0.027} & \tsecond{0.022}& \tsecond{0.056}& \tsecond{0.011}& \tfirst{0.103} & \tfirst{0.041} \\
\hline
        \end{tabular}

\vspace{0.1in}
\caption{\textbf{Detailed results on the Blender dataset \cite{mildenhall2021nerf}.} We show the PSNR, SSIM, and LPIPS (VGG) results averaged over the testing images.
 We highlight the \colorbox{tab_red}{best}, \colorbox{orange}{second}, and \colorbox{yellow}{third} values. We outperform Point-NeRF$^{col}$ \cite{xu2022pointnerf} which was evaluated with the same input point cloud as our method.
\label{tab:supp-blender}
}
\end{table*}
\begin{figure*}[ht!]
\centering
\small
\begin{tikzpicture}[
 image/.style = {text width=0.29\linewidth, 
                 inner sep=0pt, outer sep=0pt},
label/.style = { minimum height=0.4cm },
node distance = 1pt and 1pt
                        ] 
\path coordinate(last);
\node [image,below=of last,alias=last] (img00) {\includegraphics[width=\linewidth]{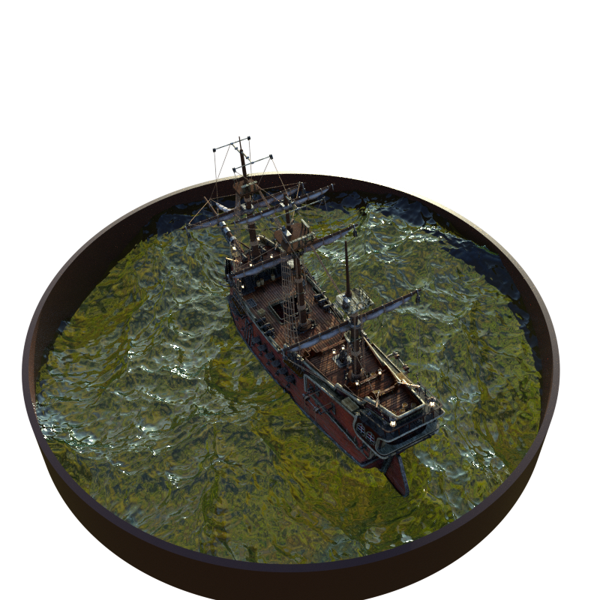}};
\draw[draw=red,line width=0.5mm] ($ (img00) + (-1.3,-1.76) $) rectangle ++(0.6,0.6);
\node [image,right=of last,alias=rlast] (img01){\includegraphics[width=\linewidth]{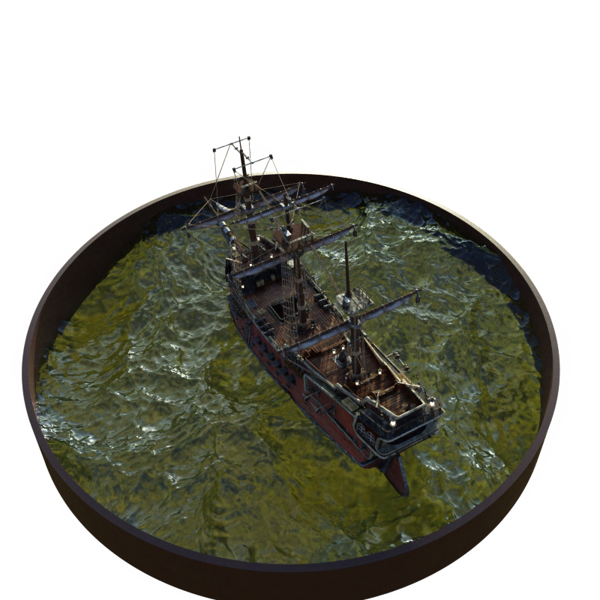}};
\draw[draw=red,line width=0.5mm] ($ (img01) + (-1.3,-1.76) $) rectangle ++(0.6,0.6);
\node [image,right=of rlast] (img02)           {\includegraphics[width=\linewidth]{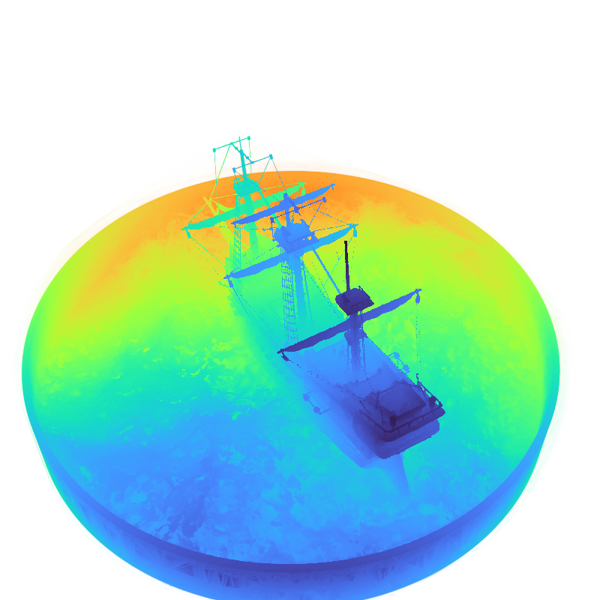}};
\draw[draw=red,line width=0.5mm] ($ (img02) + (-1.3,-1.76) $) rectangle ++(0.6,0.6);

\node [image,below=of last,alias=last] (img) {\includegraphics[width=\linewidth]{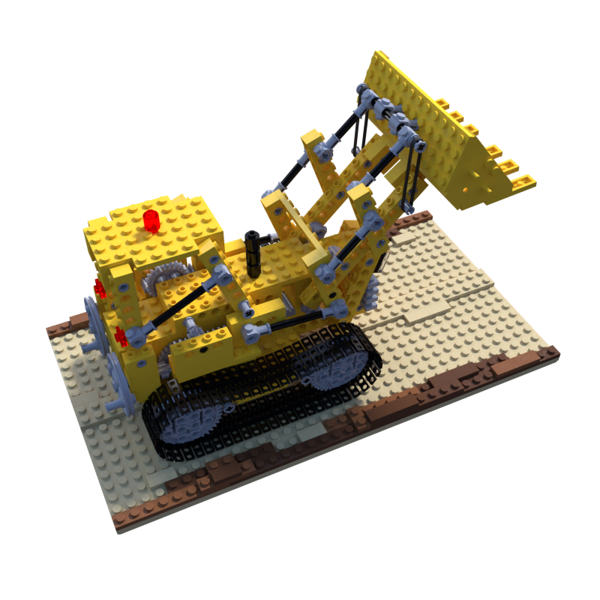}};
\draw[draw=red,line width=0.5mm] ($ (img) + (-1.6,0.4) $) rectangle ++(0.6,0.6);
\node [image,right=of last,alias=rlast] (img){\includegraphics[width=\linewidth]{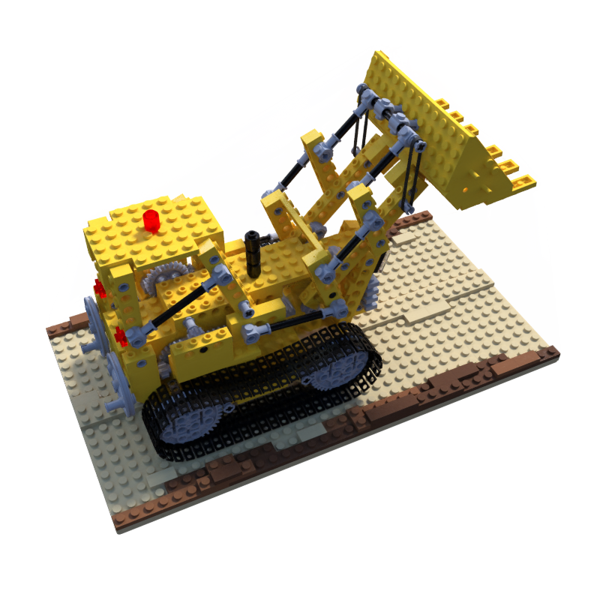}};
\draw[draw=red,line width=0.5mm] ($ (img) + (-1.6,0.4) $) rectangle ++(0.6,0.6);
\node [image,right=of rlast] (img)           {\includegraphics[width=\linewidth]{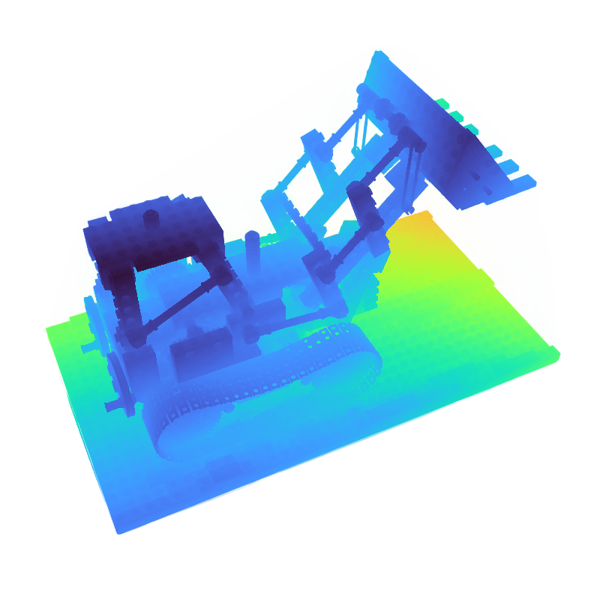}};
\draw[draw=red,line width=0.5mm] ($ (img) + (-1.6,0.4) $) rectangle ++(0.6,0.6);

\node [image,below=of last,alias=last] (img) {\includegraphics[width=\linewidth]{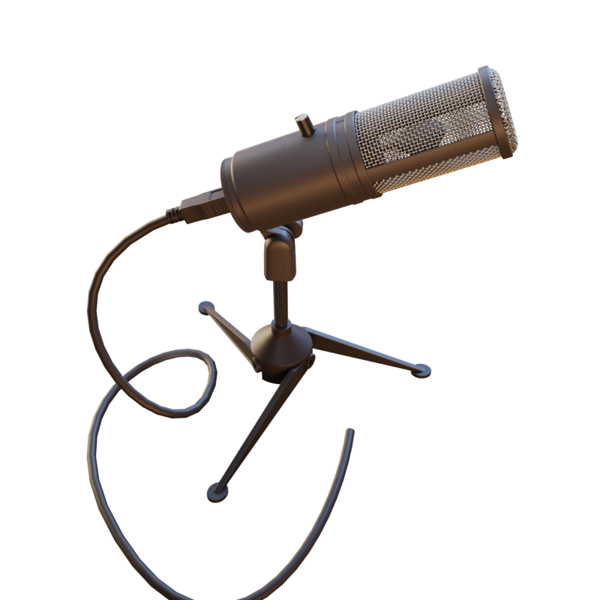}};
\draw[draw=red,line width=0.5mm] ($ (img) + (0,1.2) $) rectangle ++(0.6,0.6);
\node [image,right=of last,alias=rlast] (img){\includegraphics[width=\linewidth]{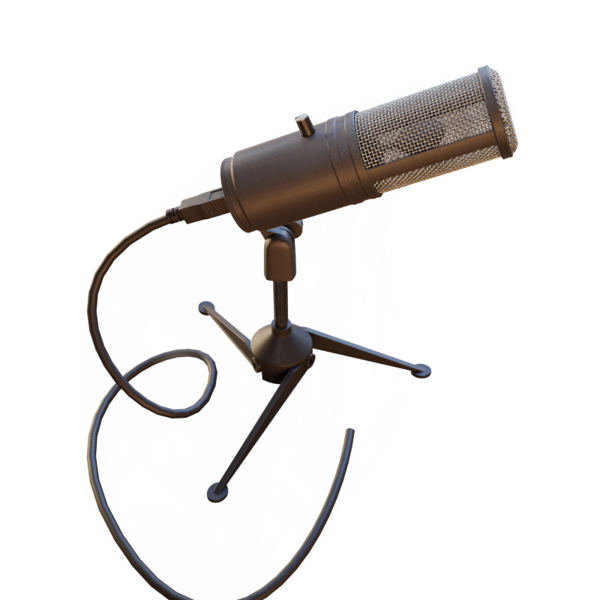}};
\draw[draw=red,line width=0.5mm] ($ (img) + (0,1.2) $) rectangle ++(0.6,0.6);
\node [image,right=of rlast] (img)           {\includegraphics[width=\linewidth]{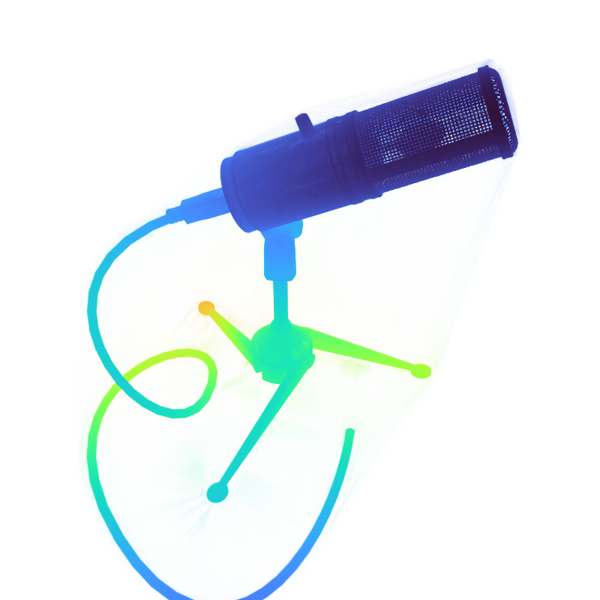}};
\draw[draw=red,line width=0.5mm] ($ (img) + (0,1.2) $) rectangle ++(0.6,0.6);

\node [image,below=of last,alias=last] (img) {\includegraphics[width=\linewidth]{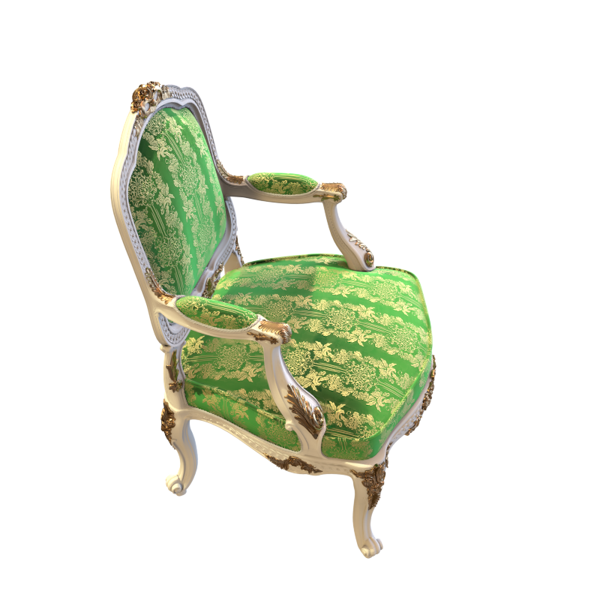}};
\draw[draw=red,line width=0.5mm] ($ (img) + (-1.35,-0.9) $) rectangle ++(0.6,0.6);
\node [image,right=of last,alias=rlast] (img){\includegraphics[width=\linewidth]{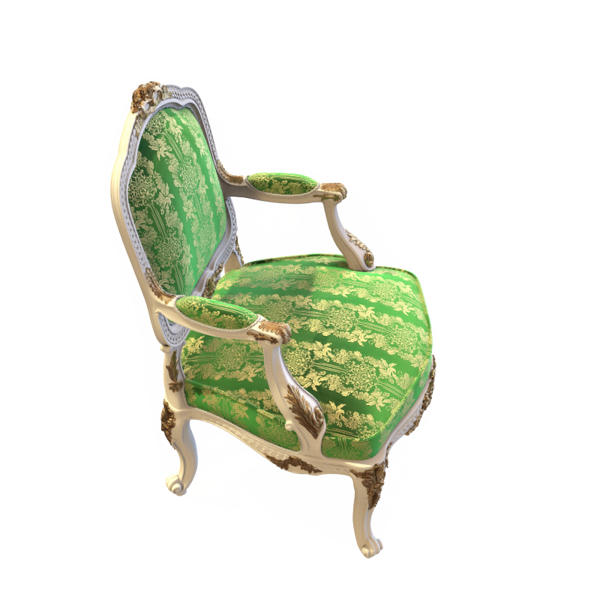}};
\draw[draw=red,line width=0.5mm] ($ (img) + (-1.35,-0.9) $) rectangle ++(0.6,0.6);
\node [image,right=of rlast] (img)           {\includegraphics[width=\linewidth]{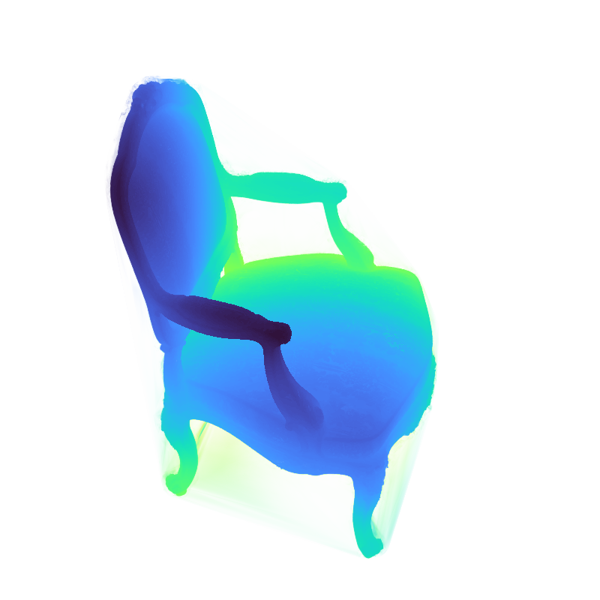}};
\draw[draw=red,line width=0.5mm] ($ (img) + (-1.35,-0.9) $) rectangle ++(0.6,0.6);
    
\node [label,above=-1pt of img00] {ground truth};
\node [label,above=-1pt of img01] {prediction};
\node [label,above=-1pt of img02] {depth};
\end{tikzpicture}

\caption{\textbf{Results on the Blender dataset \cite{mildenhall2021nerf} (part 1).} We show the \textbf{ground-truth} image, the \textbf{prediction}, and the \textbf{predicted depth map} on scenes: \textit{ship} \textbf{(top)}, \textit{lego}, \textit{mic}, and \textit{chair} \textbf{(bottom)}. The \textcolor{red}{red} squares highlight regions where there are visible errors in the images. Notice how our approach is able to recover fine textures in \textit{chair} scene and tiny details in the \textit{ship} scene geometry. However, the density seems to be non-zero in tetrahedra connecting different legs of the \textit{chair}.
\label{fig:supp-blender-1}}
\end{figure*}
\begin{figure*}[ht!]
\centering
\small
\begin{tikzpicture}[
 image/.style = {text width=0.29\linewidth, 
                 inner sep=0pt, outer sep=0pt},
label/.style = { minimum height=0.4cm },
node distance = 1pt and 1pt
                        ] 
\path coordinate(last);
\node [image,below=of last,alias=last] (img00) {\includegraphics[width=\linewidth]{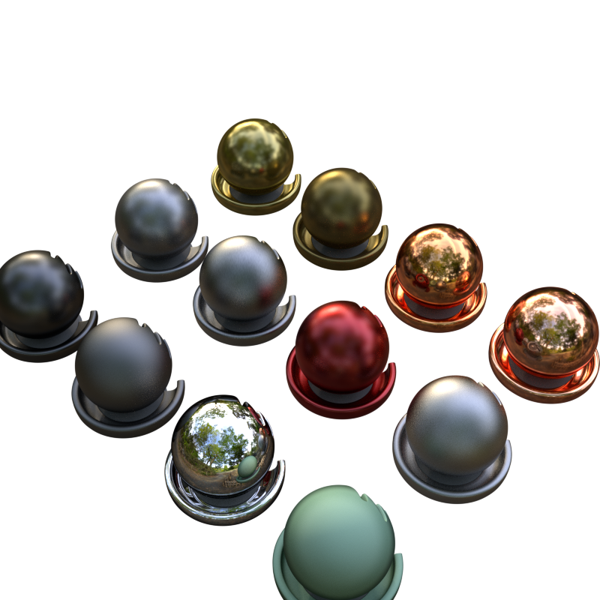}};
\draw[draw=red,line width=0.5mm] ($ (img00) + (-0.9,-1.4) $) rectangle ++(0.6,0.6);
\node [image,right=of last,alias=rlast] (img01){\includegraphics[width=\linewidth]{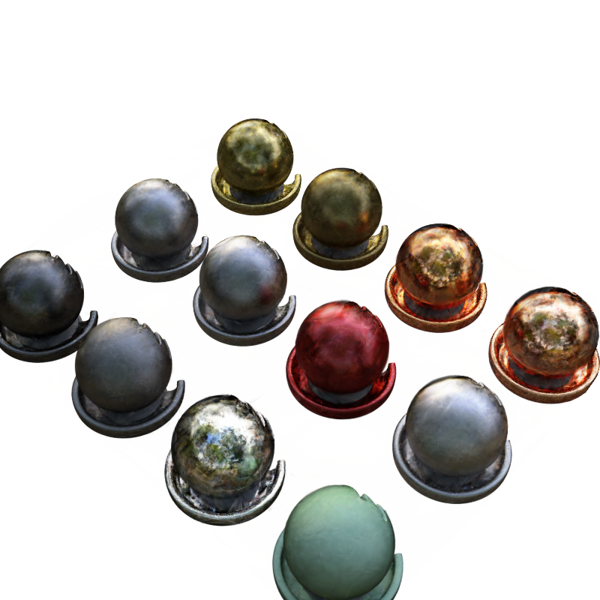}};
\draw[draw=red,line width=0.5mm] ($ (img01) + (-0.9,-1.4) $) rectangle ++(0.6,0.6);
\node [image,right=of rlast] (img02)           {\includegraphics[width=\linewidth]{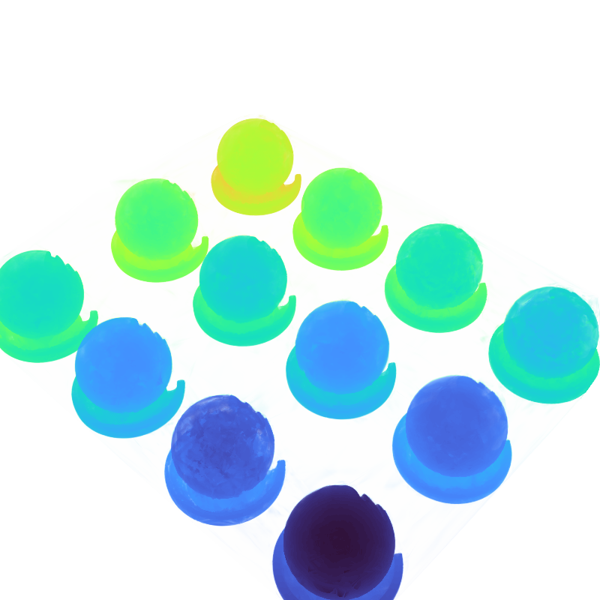}};
\draw[draw=red,line width=0.5mm] ($ (img02) + (-0.9,-1.4) $) rectangle ++(0.6,0.6);

\node [image,below=of last,alias=last] (img) {\includegraphics[width=\linewidth]{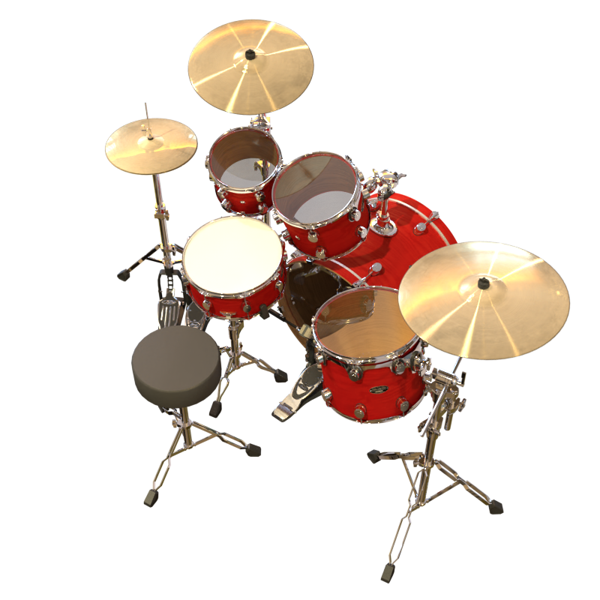}};
\draw[draw=red,line width=0.5mm] ($ (img) + (0.2,-0.4) $) rectangle ++(0.6,0.6);
\node [image,right=of last,alias=rlast] (img){\includegraphics[width=\linewidth]{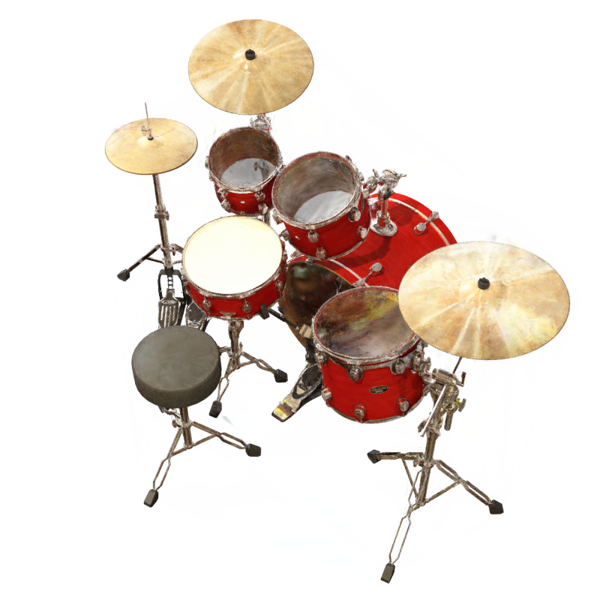}};
\draw[draw=red,line width=0.5mm] ($ (img) + (0.2,-0.4) $) rectangle ++(0.6,0.6);
\node [image,right=of rlast] (img)           {\includegraphics[width=\linewidth]{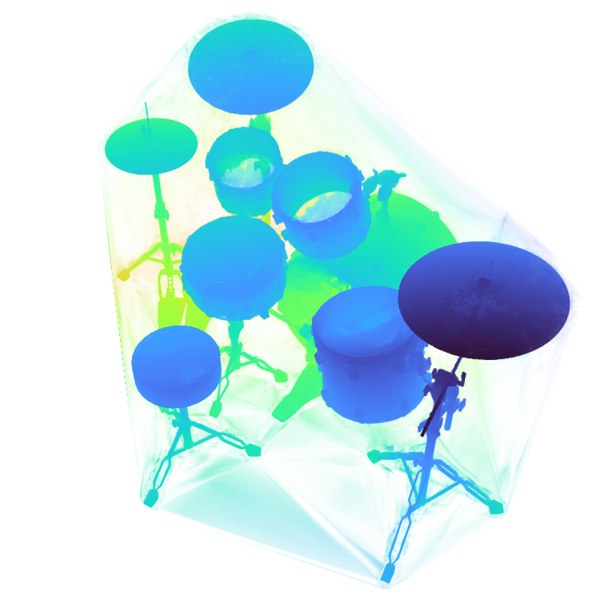}};
\draw[draw=red,line width=0.5mm] ($ (img) + (0.2,-0.4) $) rectangle ++(0.6,0.6);

\node [image,below=of last,alias=last] (img) {\includegraphics[width=\linewidth]{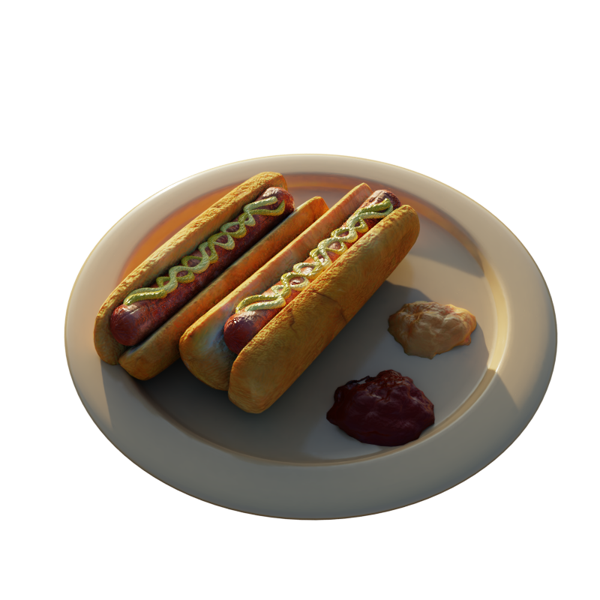}};
\draw[draw=red,line width=0.5mm] ($ (img) + (0.3,-1.1) $) rectangle ++(0.6,0.6);
\node [image,right=of last,alias=rlast] (img){\includegraphics[width=\linewidth]{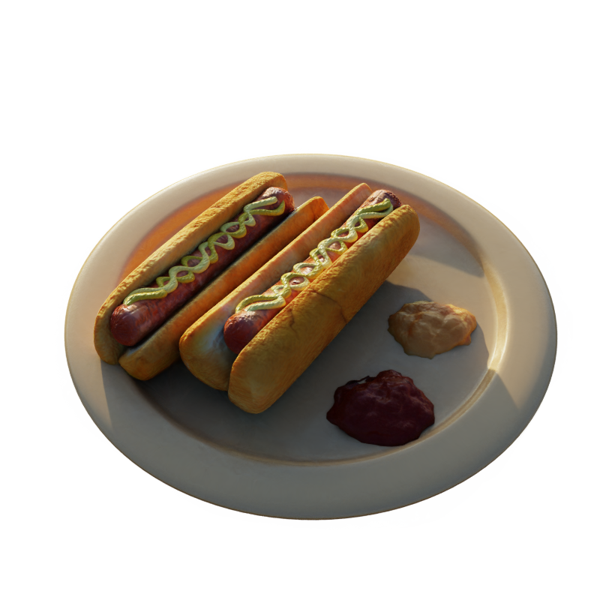}};
\draw[draw=red,line width=0.5mm] ($ (img) + (0.3,-1.1) $) rectangle ++(0.6,0.6);
\node [image,right=of rlast] (img)           {\includegraphics[width=\linewidth]{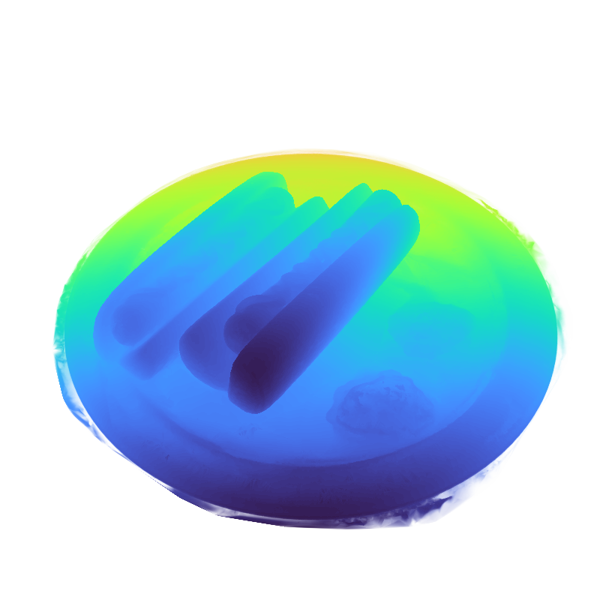}};
\draw[draw=red,line width=0.5mm] ($ (img) + (0.3,-1.1) $) rectangle ++(0.6,0.6);

\node [image,below=of last,alias=last] (img) {\includegraphics[width=\linewidth]{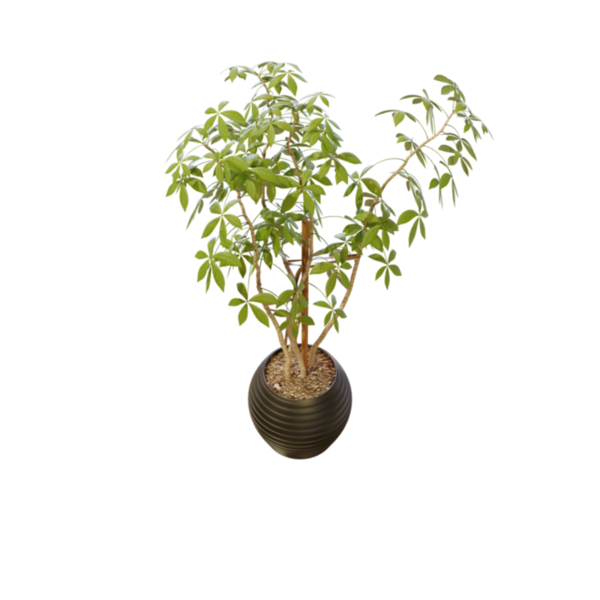}};
\draw[draw=red,line width=0.5mm] ($ (img) + (-0.3,-1.4) $) rectangle ++(0.6,0.6);
\node [image,right=of last,alias=rlast] (img){\includegraphics[width=\linewidth]{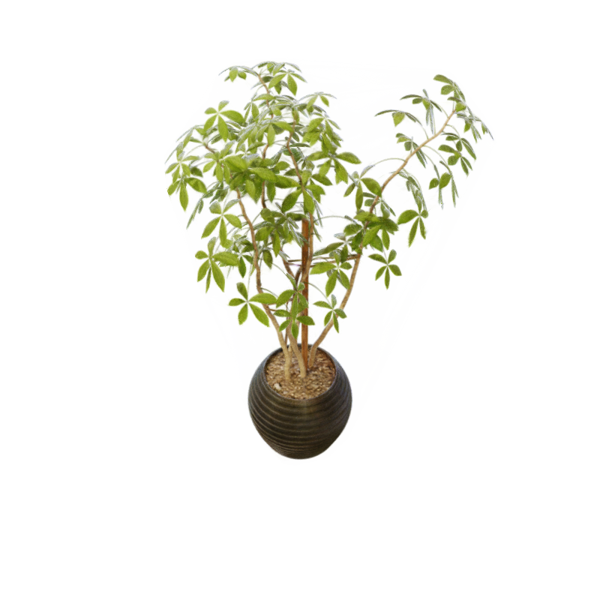}};
\draw[draw=red,line width=0.5mm] ($ (img) + (-0.3,-1.4) $) rectangle ++(0.6,0.6);
\node [image,right=of rlast] (img)           {\includegraphics[width=\linewidth]{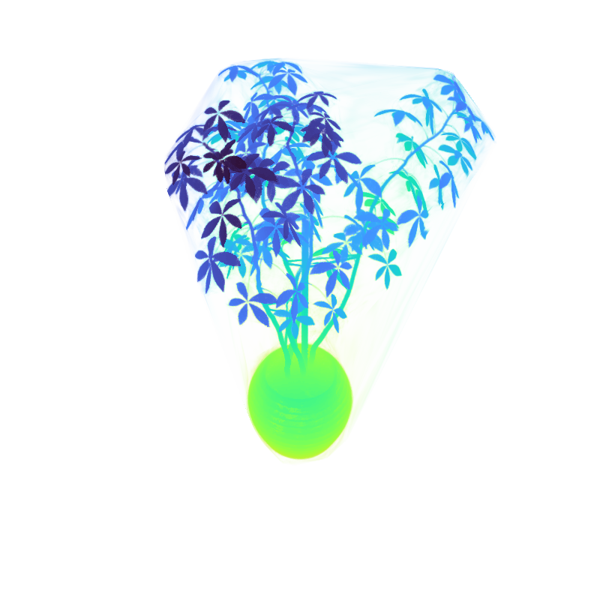}};
\draw[draw=red,line width=0.5mm] ($ (img) + (-0.3,-1.4) $) rectangle ++(0.6,0.6);

\node [label,above=-1pt of img00] {ground truth};
\node [label,above=-1pt of img01] {prediction};
\node [label,above=-1pt of img02] {depth};
\end{tikzpicture}

\caption{\textbf{Results on the Blender dataset \cite{mildenhall2021nerf} (part 2).} We show the \textbf{ground-truth} image, the \textbf{prediction}, and the \textbf{predicted depth map} on scenes: \textit{materials} \textbf{(top)}, \textit{drums}, \textit{hotdog}, and \textit{ficus} \textbf{(bottom)}. The \textcolor{red}{red} squares highlight regions where there are visible errors in the images. Some artefacts can be noticed in the top two scenes, where there are highly reflective surfaces. Also, the density seems to be non-zero in tetrahedra connecting distant parts of the 3D object.
\label{fig:supp-blender-2}}
\end{figure*}
 In Sec.~4.2 and Tab.~1 of the main paper, we presented results averaged over all scenes in the Blender dataset \cite{mildenhall2021nerf}. In the following, we present results individually per scene. The quantitative results can be seen in Table~\ref{tab:supp-blender}. We report the PSNR, SSIM, and LPIPS (VGG) \cite{zhang2018lpips} metrics. The evaluation protocol is the same as in Point-NeRF~\cite{xu2022pointnerf}, and we use the same input point cloud as Point-NeRF$^{col}$.
From the results, we can see that we outperform the closest approach to ours, Point-NeRF, in all three metrics on almost all scenes when using the same input point cloud (row Point-NeRF$^{col}$). We are slightly outperformed by Point-NeRF$^{mvs}$, which uses denser input point clouds obtained from its end-to-end optimised Multi-View Stereo (MVS) pipeline. The results are more pronounced on the \textit{ficus} scene where Point-NeRF performs exceptionally well, outperforming other baselines, including Mip-NeRF \cite{barron2021mipnerf}.
Note that both Point-NeRF configurations grow the point cloud during training and, therefore, the complexity of the scene representation grows. For us, the points are fixed, and the number of parameters stays the same. 
In most scenes, we also outperform Plenoxels \cite{fridovich2022plenoxels} which uses a sparse grid. Note that same as Point-NeRF, Plenoxels also gradually increases the representation complexity by subdividing the grid resolution at predefined training epochs. 
Even though Mip-NeRF \cite{barron2021mipnerf} outperforms our approach in terms of PSNR, our method is slightly better in terms of SSIM and on par with Mip-NeRF \cite{barron2021mipnerf} in terms of LPIPS.

We also show rendered images from all Blender scenes in Figures~\ref{fig:supp-blender-1} and \ref{fig:supp-blender-2}. Some artefacts can only be noticed on scenes with highly reflective surfaces -- \textit{materials}, and \textit{drums}. By closely inspecting the produced depth maps, one can notice that sometimes the density is non-zero in large tetrahedra connecting different parts of the object. A possible cause could be the combination of the training process and the implicit bias of our model. Since the tetrahedra field uses barycentric interpolation, the features will change linearly in the tetrahedra connecting different parts of the object. However, these tetrahedra should have a density of zero everywhere except for the regions close to the vertices. For the shallow MLP, it is difficult to represent such a function, and there is not enough pressure in the optimisation process to enforce it because the error will be close to zero since the background is white, without any texture, independently of how the density is distributed in these regions.

\clearpage
\section{Tanks and Temples results}\label{sec:tt}
We show the detailed per-scene results for the Tanks and Temples dataset \cite{knapitsch2017tanks}. Note that to be able to compare with the Point-NeRF method \cite{xu2022pointnerf}, we used the data pre-processing and splits proposed by NSVF \cite{liu2020nsvf}, where the background is masked out.
The quantitative results can be seen in Table~\ref{tab:supp-tt}.
We report the PSNR, SSIM, and LPIPS (Alex) \cite{zhang2018lpips} metrics. The evaluation protocol is the same as in Point-NeRF~\cite{xu2022pointnerf}, but we evaluate on the original resolution, the same as other compared methods. To be able to compare with Point-NeRF~\cite{xu2022pointnerf} which evaluated on a lower-resolution images\footnote{\url{https://github.com/Xharlie/pointnerf/issues/62}}, we have recomputed its metrics with the same full image resolution $1920\times1080$. To obtain the initial point clouds, we used the dense COLMAP reconstruction, which was computed using the known intrinsic and extrinsic camera parameters. However, the NSVF~\cite{liu2020nsvf} published split had corrupted camera parameters for the \textit{ignatius} scene, and we had to run COLMAP reconstruction from scratch to obtain both the camera poses and intrinsics. This is likely the reason for the worse results on that scene. Otherwise, we outperform all baseline methods on all metrics.

\begin{table*}[b]
    \small
    \newcommand{\tfirst}[1]{\cellcolor{tab_red} #1}
    \newcommand{\tsecond}[1]{\cellcolor{orange} #1}
    \newcommand{\tthird}[1]{\cellcolor{yellow} #1}
    \setlength\tabcolsep{6pt}
      \centering
        \begin{tabular}{@{}l|ccccc|c}
\multicolumn{1}{c}{\,} & \multicolumn{6}{c}{PSNR~$\uparrow$} \\
        \multicolumn{1}{l|}{} & \textit{barn} & \textit{caterpillar} & \textit{family} & \textit{ignatius}            & \textit{truck}     & \textit{mean}                 \\ \hline
NV~\cite{lombardi2019neural}         & 20.82          & 20.71          & 28.72          & 26.54          & 21.71          & 23.70          \\
NeRF~\cite{mildenhall2021nerf}       & 24.05          & 23.75          & 30.29          & 25.43          & 25.36          & 25.78          \\
NSVF~\cite{liu2020nsvf}              & \tthird{27.16} & \tsecond{26.44}& \tsecond{33.58}& \tsecond{27.91}& \tsecond{26.92}& \tsecond{28.40}\\
Point-NeRF~\cite{xu2022pointnerf}$^*$& \tsecond{27.40}& \tthird{25.58} & \tthird{33.57} & \tfirst{28.39} & \tthird{26.83} & \tthird{28.35} \\
\textbf{\ours}             & \tfirst{28.86} & \tfirst{26.64} & \tfirst{34.27} & \tthird{27.17}$^\dagger$\hspace{-4pt} & \tfirst{27.58} & \tfirst{28.90} \\\hline

\multicolumn{7}{c}{} \\
\multicolumn{1}{c}{\,} & \multicolumn{6}{c}{SSIM~$\uparrow$} \\
        \multicolumn{1}{l|}{} & \textit{barn} & \textit{caterpillar} & \textit{family} & \textit{ignatius}            & \textit{truck}     & \textit{mean}                 \\ \hline

NV~\cite{lombardi2019neural}         & 0.721          & 0.819          & 0.916          & \tfirst{0.992} & 0.793          & 0.848          \\
NeRF~\cite{mildenhall2021nerf}       & 0.750          & 0.860          & 0.932          & 0.920          & 0.860          & 0.864          \\
NSVF~\cite{liu2020nsvf}              & \tthird{0.823} & \tthird{0.900} & \tthird{0.954} & 0.930          & \tthird{0.895} & \tthird{0.900} \\
Point-NeRF~\cite{xu2022pointnerf}$^*$& \tsecond{0.908}& \tsecond{0.927}& \tsecond{0.976}& \tthird{0.959} & \tsecond{0.939}& \tsecond{0.942}\\
\textbf{\ours}             & \tfirst{0.942} & \tfirst{0.944} & \tfirst{0.985} & \tsecond{0.962}& \tfirst{0.952} & \tfirst{0.957} \\\hline

\multicolumn{7}{c}{} \\
\multicolumn{1}{c}{\,} & \multicolumn{6}{c}{LPIPS~$\downarrow$} \\
        \multicolumn{1}{l|}{} & \textit{barn} & \textit{caterpillar} & \textit{family} & \textit{ignatius}            & \textit{truck}     & \textit{mean}                 \\ \hline
NV~\cite{lombardi2019neural}         & 0.117          & 0.312          & 0.479          & 0.280          & 0.111          & 0.260          \\
NeRF~\cite{mildenhall2021nerf}       & \tthird{0.111} & 0.192          & 0.395          & 0.196          & 0.098          & 0.198          \\
NSVF~\cite{liu2020nsvf}              & \tsecond{0.106}& \tthird{0.148} & \tthird{0.307} & \tthird{0.141} & \tsecond{0.063}& \tthird{0.153} \\
Point-NeRF~\cite{xu2022pointnerf}$^*$& 0.142          & \tsecond{0.118}& \tsecond{0.034}& \tsecond{0.064}& \tthird{0.091} & \tsecond{0.090}\\
\textbf{\ours}                      & \tfirst{0.087} & \tfirst{0.077} & \tfirst{0.021} & \tfirst{0.050} & \tfirst{0.062} & \tfirst{0.059} \\\hline

\end{tabular}

\vspace{0.1in}
\caption{\textbf{Tanks and Temples results.}
We show the PSNR, SSIM, and LPIPS (Alex) results averaged over the testing images.
 We highlight the \colorbox{tab_red}{best}, \colorbox{orange}{second}, and \colorbox{yellow}{third} values. We outperform all compared methods in all metrics on all scenes except for \textit{ignatius}, where we did not have the correct camera parameters and had to run camera pose estimation prior to training.
\label{tab:supp-tt}
}
\end{table*}
\begin{figure*}[ht!]
\centering
\small
\begin{tikzpicture}[
 image/.style = {text width=0.23\linewidth, 
                 inner sep=0pt, outer sep=0pt},
label/.style = { minimum height=0.4cm },
node distance = 1pt and 1pt
                        ] 
\path coordinate(last);

\node [image,below=of last,alias=last] (img00) {\includegraphics[width=\linewidth]{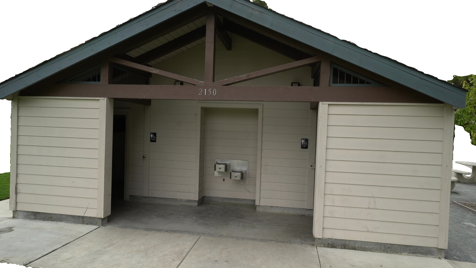}};
\draw[draw=red,line width=0.5mm] ($ (img00) + (-0.5,0.5) $) rectangle ++(0.6,0.6);
\node [image,right=of last,alias=rlast] (img01){\includegraphics[width=\linewidth]{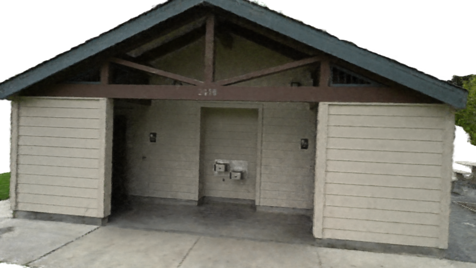}};
\draw[draw=red,line width=0.5mm] ($ (img01) + (-0.5,0.5) $) rectangle ++(0.6,0.6);
\node [image,right=of rlast,alias=rlast](img02){\includegraphics[width=\linewidth]{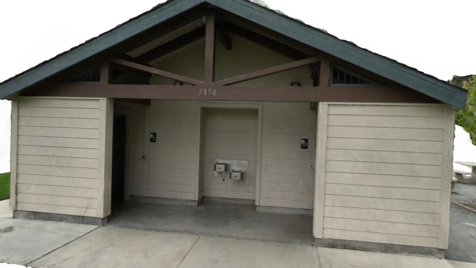}};
\draw[draw=red,line width=0.5mm] ($ (img02) + (-0.5,0.5) $) rectangle ++(0.6,0.6);
\node [image,right=of rlast] (img03)           {\includegraphics[width=\linewidth]{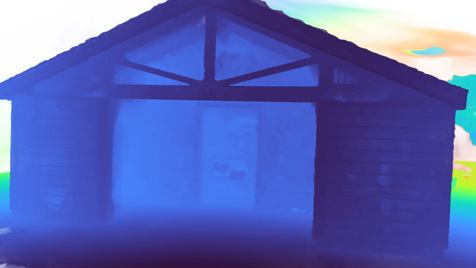}};
\draw[draw=red,line width=0.5mm] ($ (img03) + (-0.5,0.5) $) rectangle ++(0.6,0.6);

\node [image,below=of last,alias=last] (img) {\includegraphics[width=\linewidth]{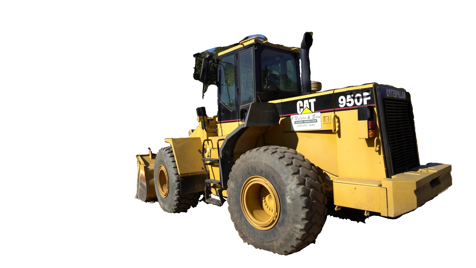}};
\draw[draw=red,line width=0.5mm] ($ (img) + (1.0,-0.8) $) rectangle ++(0.6,0.6);
\node [image,right=of last,alias=rlast] (img){\includegraphics[width=\linewidth]{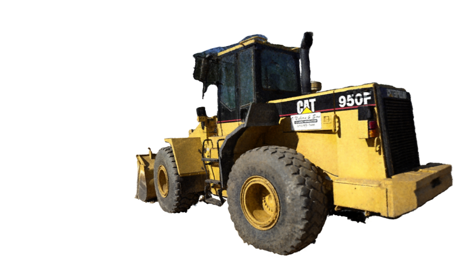}};
\draw[draw=red,line width=0.5mm] ($ (img) + (1.0,-0.8) $) rectangle ++(0.6,0.6);
\node [image,right=of rlast,alias=rlast](img){\includegraphics[width=\linewidth]{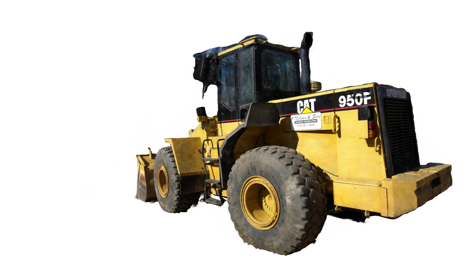}};
\draw[draw=red,line width=0.5mm] ($ (img) + (1.0,-0.8) $) rectangle ++(0.6,0.6);
\node [image,right=of rlast] (img)           {\includegraphics[width=\linewidth]{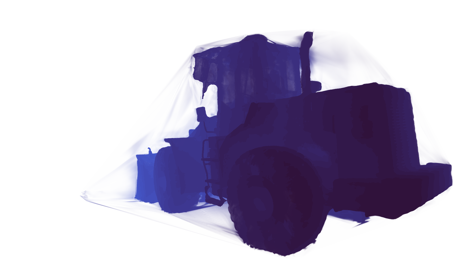}};
\draw[draw=red,line width=0.5mm] ($ (img) + (1.0,-0.8) $) rectangle ++(0.6,0.6);

\node [image,below=of last,alias=last] (img) {\includegraphics[width=\linewidth]{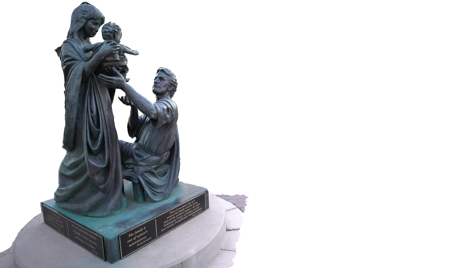}};
\node [image,right=of last,alias=rlast] (img){\includegraphics[width=\linewidth]{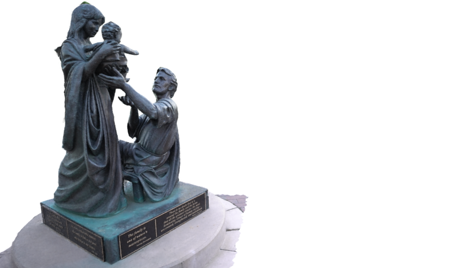}};
\node [image,right=of rlast,alias=rlast](img){\includegraphics[width=\linewidth]{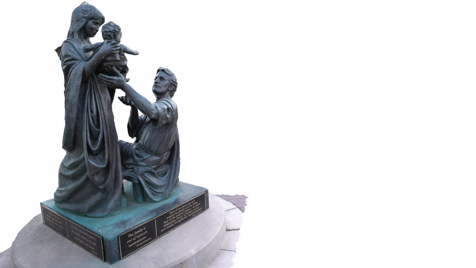}};
\node [image,right=of rlast] (img)           {\includegraphics[width=\linewidth]{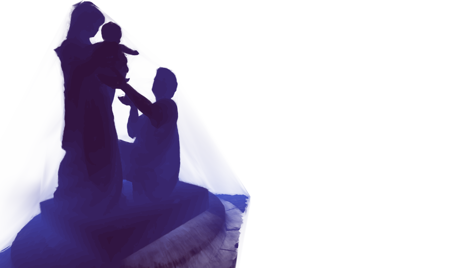}};

\node [image,below=of last,alias=last] (img) {\includegraphics[width=\linewidth]{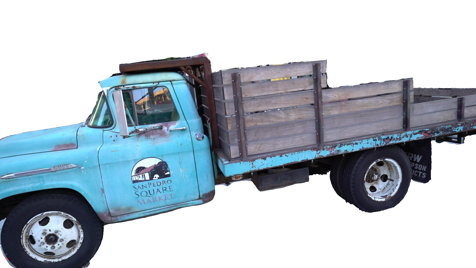}};
\draw[draw=red,line width=0.5mm] ($ (img) + (-1.9,-1.14) $) rectangle ++(0.6,0.6);
\node [image,right=of last,alias=rlast] (img){\includegraphics[width=\linewidth]{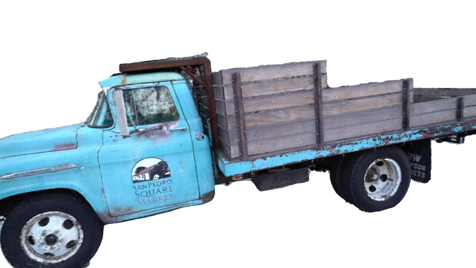}};
\draw[draw=red,line width=0.5mm] ($ (img) + (-1.9,-1.14) $) rectangle ++(0.6,0.6);
\node [image,right=of rlast,alias=rlast](img){\includegraphics[width=\linewidth]{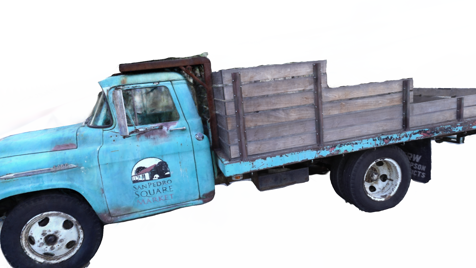}};
\draw[draw=red,line width=0.5mm] ($ (img) + (-1.9,-1.14) $) rectangle ++(0.6,0.6);
\node [image,right=of rlast] (img)           {\includegraphics[width=\linewidth]{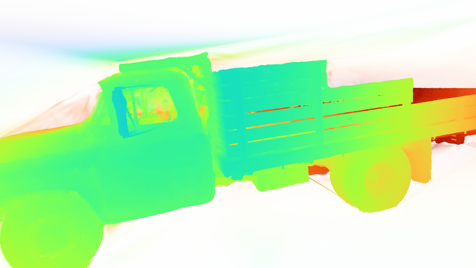}};
\draw[draw=red,line width=0.5mm] ($ (img) + (-1.9,-1.14) $) rectangle ++(0.6,0.6);

\node [image,below=of last,alias=last] (img) {\includegraphics[width=\linewidth]{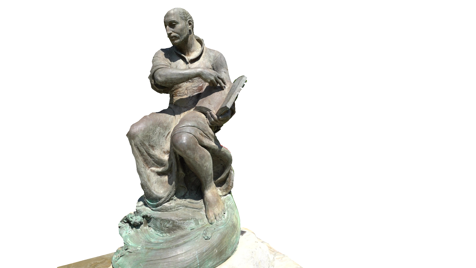}};
\draw[draw=red,line width=0.5mm] ($ (img) + (-0.9,0.1) $) rectangle ++(0.6,0.6);
\node [image,right=of last,alias=rlast] (img){\includegraphics[width=\linewidth]{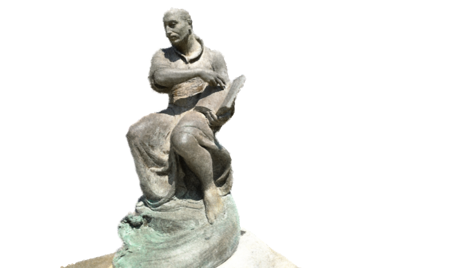}};
\draw[draw=red,line width=0.5mm] ($ (img) + (-0.9,0.1) $) rectangle ++(0.6,0.6);
\node [image,right=of rlast,alias=rlast](img){\includegraphics[width=\linewidth]{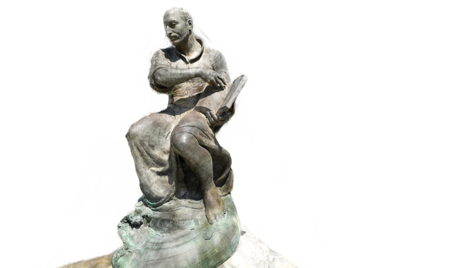}};
\draw[draw=red,line width=0.5mm] ($ (img) + (-0.9,0.1) $) rectangle ++(0.6,0.6);
\node [image,right=of rlast] (img)           {\includegraphics[width=\linewidth]{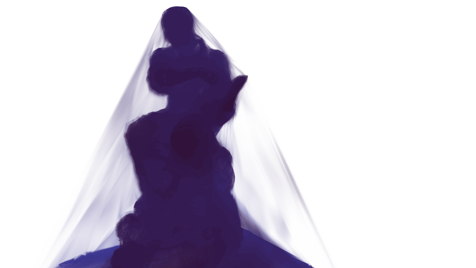}};
\draw[draw=red,line width=0.5mm] ($ (img) + (-0.9,0.1) $) rectangle ++(0.6,0.6);

\node [label,above=-1pt of img00] {ground truth};
\node [label,above=-1pt of img01] {Point-NeRF};
\node [label,above=-1pt of img02] {\textbf{Tetra-NeRF}};
\node [label,above=-1pt of img03] {\textbf{Tetra-NeRF} depth};
\end{tikzpicture}

\caption{\textbf{Results on the Tanks and Temples dataset \cite{knapitsch2017tanks}.}
 We show the \textbf{ground-truth} image, the \textbf{prediction}, and the \textbf{predicted depth map} on scenes: \textit{barn} \textbf{(top)}, \textit{caterpillar}, \textit{family}, \textit{truck}, and \textit{ignatius} \textbf{(bottom)}. 
 We also show a comparison with Point-NeRF~\cite{xu2022pointnerf}.
 The \textcolor{red}{red} squares highlight regions where our method has fewer artefacts compared to Point-NeRF.
 Notice how our method produces less noisy images compared to Point-NeRF. We can observe similar non-zero density artefacts to the ones observed on the Blender dataset. Again, we attribute it to a combination of implicit bias and non-textured background. 
\label{fig:supp-tt}
}
\end{figure*}

We also extend Fig.~6 from the main paper and show more qualitative results in Figure~\ref{fig:supp-tt}.
Compared to Point-NeRF~\cite{xu2022pointnerf}, our method produces less noisy images. When looking at the depth maps, we can observe similar non-zero density artefacts to the ones observed on the Blender dataset. Same as in the Blender dataset case, a likely cause could be a combination of the implicit bias of our method and the usage of non-textured background.

\clearpage
\section{Mip-NeRF 360 results}\label{sec:mipnerf360}
\begin{table*}[b!]
    \newcommand{\tfirst}[1]{\cellcolor{tab_red} #1}
    \newcommand{\tsecond}[1]{\cellcolor{orange} #1}
    \newcommand{\tthird}[1]{\cellcolor{yellow} #1}
    \centering
    \small
    \begin{tabular}{@{}l|ccccc|cccc}
    \multicolumn{1}{c}{\,} & \multicolumn{9}{c}{PSNR} \\
    & \multicolumn{5}{c|}{Outdoor} & \multicolumn{4}{c}{Indoor} \\
 & \textit{bicycle}& \textit{flowers}& \textit{garden}& \textit{stump}& \textit{treehill}& \textit{room}& \textit{counter}& \textit{kitchen}& \textit{bonsai}\\ \hline 
NeRF~\cite{mildenhall2021nerf, jaxnerf2020github}   & 21.76          & 19.40          & 23.11          & 21.73          & 21.28          & 28.56          & 25.67          & 26.31          & 26.81          \\
mip-NeRF~\cite{barron2021mipnerf}                   & 21.69          & 19.31          & 23.16          & 23.10          & 21.21          & 28.73          & 25.59          & 26.47          & 27.13          \\
NeRF++~\cite{zhang2020nerf++}                          & 22.64          & \tthird{20.31} & 24.32          & 24.34          & \tfirst{22.20} & \tthird{28.87} & 26.38          & 27.80          & \tthird{29.15} \\
Deep Blending~\cite{hedman2018deep}                 & 21.09          & 18.13          & 23.61          & 24.08          & 20.80          & 27.20          & 26.28          & 25.02          & 27.08          \\
Point-Based Neural Rendering~\cite{kopanas2021point}& 21.64          & 19.28          & 22.50          & 23.90          & 20.98          & 26.99          & 25.23          & 24.47          & 28.42          \\
Stable View Synthesis~\cite{riegler2021stable}         & \tthird{22.79} & 20.15          & \tsecond{25.99}& \tthird{24.39} & \tthird{21.72} & \tsecond{28.93}& \tthird{26.40} & \tthird{28.49} & 29.07          \\
mip-NeRF 360~\cite{barron2022mipnerf360}                                       & \tfirst{23.95} & \tfirst{21.60} & \tthird{25.09} & \tfirst{25.98} & \tsecond{21.99}& 28.24          & \tfirst{28.40} & \tfirst{30.81} & \tsecond{30.27}\\
\textbf{\ours}                          & \tsecond{23.53}& \tsecond{20.36}& \tfirst{26.15} & \tsecond{24.42}& 21.41          & \tfirst{32.02} & \tsecond{28.02}& \tsecond{29.66}& \tfirst{31.13} \\\hline

    \multicolumn{10}{c}{} \\
    \multicolumn{1}{c}{\,} & \multicolumn{9}{c}{SSIM} \\
    & \multicolumn{5}{c|}{Outdoor} & \multicolumn{4}{c}{Indoor} \\
 & \textit{bicycle}& \textit{flowers}& \textit{garden}& \textit{stump}& \textit{treehill}& \textit{room}& \textit{counter}& \textit{kitchen}& \textit{bonsai}\\ \hline 
NeRF~\cite{mildenhall2021nerf,jaxnerf2020github}    & 0.455          & 0.376          & 0.546          & 0.453          & 0.459          & 0.843          & 0.775          & 0.749          & 0.792          \\
mip-NeRF~\cite{barron2021mipnerf}                   & 0.454          & 0.373          & 0.543          & 0.517          & 0.466          & 0.851          & 0.779          & 0.745          & 0.818          \\
NeRF++~\cite{zhang2020nerf++}                          & 0.526          & 0.453          & 0.635          & 0.594          & 0.530          & 0.852          & 0.802          & 0.816          & 0.876          \\
Deep Blending~\cite{hedman2018deep}                 & 0.466          & 0.320          & 0.675          & 0.634          & 0.523          & 0.868          & 0.856          & 0.768          & 0.883          \\
Point-Based Neural Rendering~\cite{kopanas2021point}& 0.608          & \tthird{0.487} & 0.735          & \tthird{0.651} & \tthird{0.579} & 0.887          & \tthird{0.868} & 0.876          & \tthird{0.919} \\
Stable View Synthesis~\cite{riegler2021stable}         & \tsecond{0.663}& \tsecond{0.541}& \tfirst{0.818} & \tsecond{0.683}& \tsecond{0.606}& \tsecond{0.905}& \tsecond{0.886}& \tsecond{0.910}& \tsecond{0.925}\\
mip-NeRF 360~\cite{barron2022mipnerf360}                                        & \tfirst{0.687} & \tfirst{0.582} & \tsecond{0.800}& \tfirst{0.745} & \tfirst{0.619} & \tfirst{0.907} & \tfirst{0.890} & \tfirst{0.916} & \tfirst{0.932} \\
\textbf{\ours}                                      & \tthird{0.614} & 0.470          & \tthird{0.775} & 0.613          & 0.456          & \tthird{0.894} & 0.850          & \tthird{0.877} & 0.905          \\\hline

    \multicolumn{10}{c}{} \\
    \multicolumn{1}{c}{\,} & \multicolumn{9}{c}{LPIPS} \\
     & \multicolumn{5}{c|}{Outdoor} & \multicolumn{4}{c}{Indoor} \\
 & \textit{bicycle}& \textit{flowers}& \textit{garden}& \textit{stump}& \textit{treehill}& \textit{room}& \textit{counter}& \textit{kitchen}& \textit{bonsai}\\ \hline 
NeRF \cite{mildenhall2021nerf, jaxnerf2020github}   & 0.536          & 0.529          & 0.415          & 0.551          & 0.546          & 0.353          & 0.394          & 0.335          & 0.398          \\
mip-NeRF \cite{barron2021mipnerf}                   & 0.541          & 0.535          & 0.422          & 0.490          & 0.538          & 0.346          & 0.390          & 0.336          & 0.370          \\
NeRF++ \cite{zhang2020nerf++}                          & 0.455          & 0.466          & 0.331          & 0.416          & 0.466          & 0.335          & 0.351          & 0.260          & 0.291          \\
Deep Blending~\cite{hedman2018deep}                 & 0.377          & 0.476          & 0.231          & 0.351          & 0.383          & 0.266          & 0.258          & 0.246          & 0.275          \\
Point-Based Neural Rendering~\cite{kopanas2021point}& 0.313          & \tthird{0.372} & 0.197          & 0.303          & \tsecond{0.325}& 0.216          & 0.209          & 0.160          & \tthird{0.178} \\
Stable View Synthesis~\cite{riegler2021stable}         & \tfirst{0.243} & \tfirst{0.317} & \tsecond{0.137}& \tthird{0.281} & \tfirst{0.286} & \tsecond{0.182}& \tsecond{0.168}& \tsecond{0.125}& \tsecond{0.164}\\
mip-NeRF 360~\cite{barron2022mipnerf360}                                        & \tthird{0.296} & \tsecond{0.343}& \tthird{0.173} & \tfirst{0.258} & \tthird{0.338} & \tthird{0.208} & \tthird{0.206} & \tthird{0.129} & 0.182          \\
\textbf{\ours}                          & \tsecond{0.271}& 0.378          & \tfirst{0.136} & \tsecond{0.274}& 0.429          & \tfirst{0.104} & \tfirst{0.127} & \tfirst{0.098} & \tfirst{0.084} \\\hline
    \end{tabular}

\vspace{0.1in}
\caption{\textbf{Detailed Mip-NeRF 360 \cite{barron2022mipnerf360} results.}
 We show the PSNR, SSIM, and LPIPS (Alex) results averaged over the testing images.
 We highlight the \colorbox{tab_red}{best}, \colorbox{orange}{second}, and \colorbox{yellow}{third} values.
Our method has a worse SSIM. However, \ours seems to perform comparably to Mip-NeRF 360 \cite{barron2022mipnerf360} on most scenes in terms of PSNR and LPIPS, where we seem to achieve a slightly higher LPIPS and a slightly lower PSNR. We also outperform Stable View Synthesis \cite{riegler2021stable} and all competitors other than Mip-NeRF~360 on all scenes in terms of PSNR.
\label{tab:supp-mipnerf360}
}
\end{table*}
\begin{figure*}[ht!]
\centering
\small
\begin{tikzpicture}[
 image/.style = {text width=0.31\linewidth, 
                 inner sep=0pt, outer sep=0pt},
label/.style = { minimum height=0.4cm },
node distance = 1pt and 1pt
                        ] 
\path coordinate(last);
    \node [image,below=of last,alias=last] (img00)
    {\includegraphics[width=\linewidth]{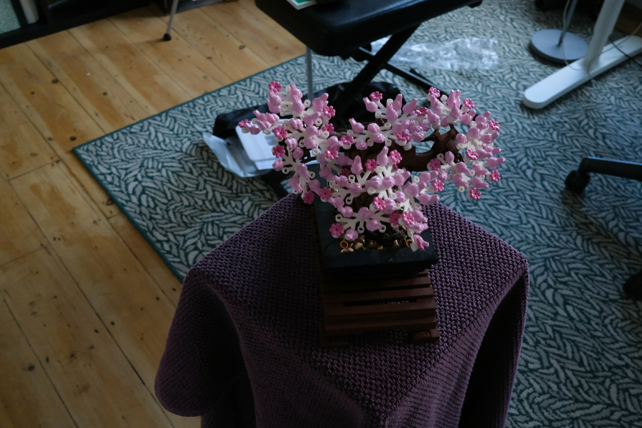}};
\draw[draw=red,line width=0.5mm] ($ (img00) + (-1.8,-1.76) $) rectangle ++(0.6,0.6);
\node [image,right=of last,alias=rlast] (img01)
    {\includegraphics[width=\linewidth]{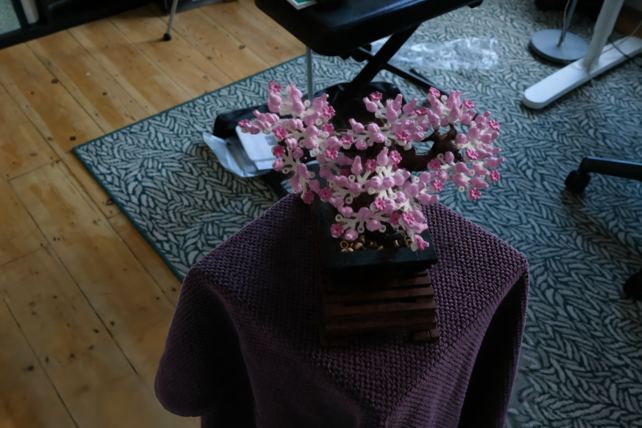}};
\draw[draw=red,line width=0.5mm] ($ (img01) + (-1.8,-1.76) $) rectangle ++(0.6,0.6);
\node [image,right=of rlast] (img02) 
    {\includegraphics[width=\linewidth]{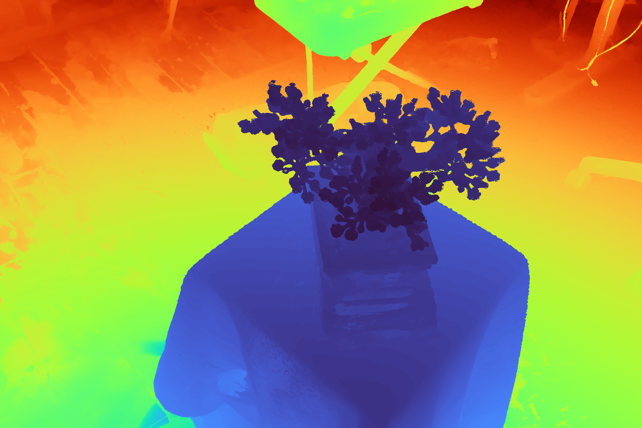}};
\draw[draw=red,line width=0.5mm] ($ (img02) + (-1.8,-1.76) $) rectangle ++(0.6,0.6);

    \node [image,below=of last,alias=last] (img)
    {\includegraphics[width=\linewidth]{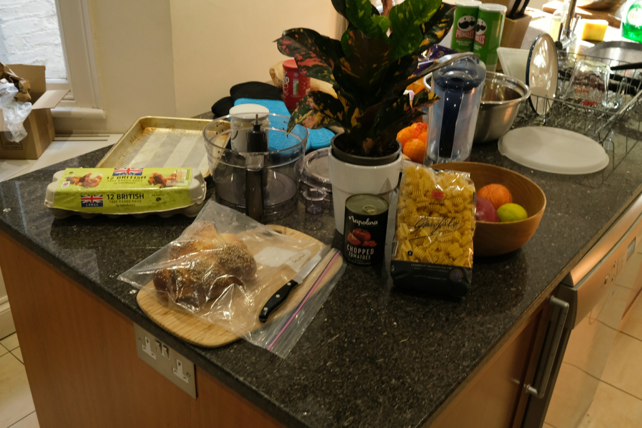}};
    \draw[draw=red,line width=0.5mm] ($ (img) + (1.6,0.8) $) rectangle ++(0.6,0.6);
\node [image,right=of last,alias=rlast] (img)
    {\includegraphics[width=\linewidth]{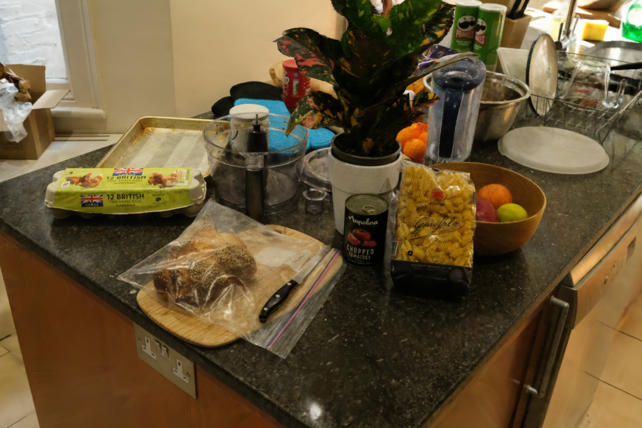}};
    \draw[draw=red,line width=0.5mm] ($ (img) + (1.6,0.8) $) rectangle ++(0.6,0.6);
\node [image,right=of rlast] (img) 
    {\includegraphics[width=\linewidth]{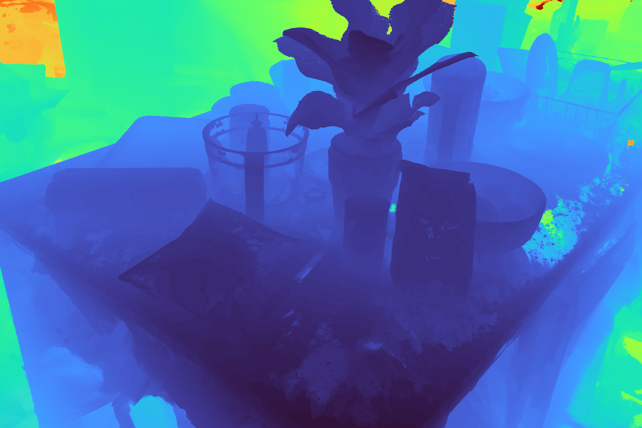}};
    \draw[draw=red,line width=0.5mm] ($ (img) + (1.6,0.8) $) rectangle ++(0.6,0.6);
    
\node [image,below=of last,alias=last] (img) {\includegraphics[width=\linewidth]{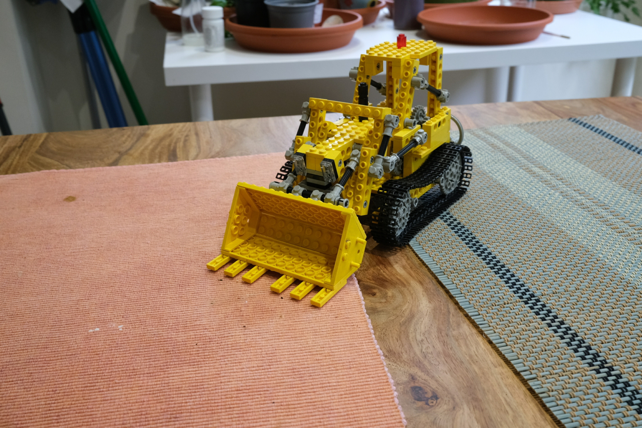}};
\draw[draw=red,line width=0.5mm] ($ (img) + (2.08,1.18) $) rectangle ++(0.6,0.6);
\node [image,right=of last,alias=rlast] (img){\includegraphics[width=\linewidth]{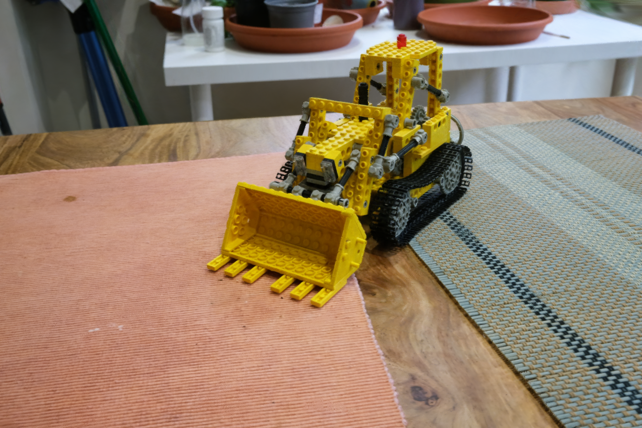}};
\draw[draw=red,line width=0.5mm] ($ (img) + (2.08,1.18) $) rectangle ++(0.6,0.6);
\node [image,right=of rlast] (img)           {\includegraphics[width=\linewidth]{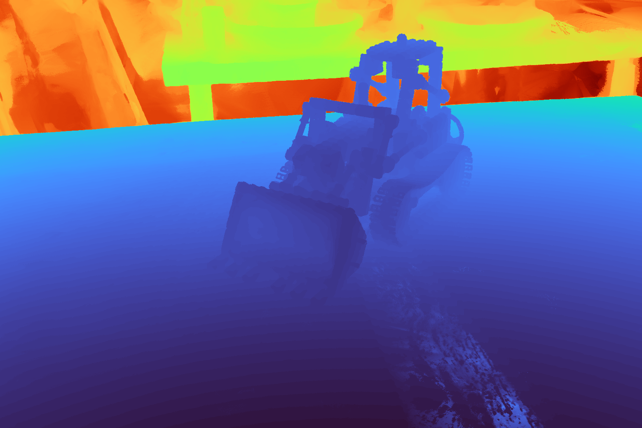}};
\draw[draw=red,line width=0.5mm] ($ (img) + (2.08,1.18) $) rectangle ++(0.6,0.6);

\node [image,below=of last,alias=last] (img) {\includegraphics[width=\linewidth]{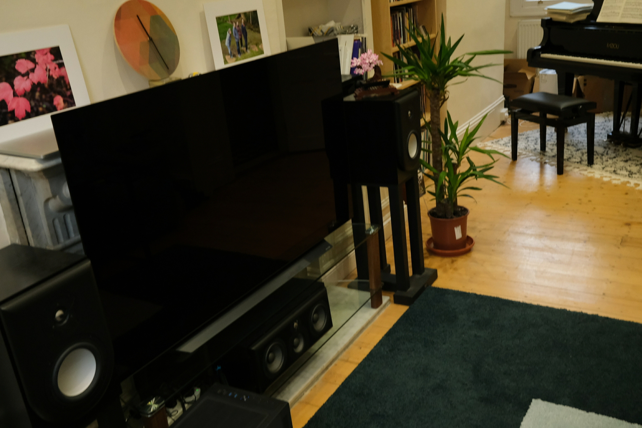}};
\draw[draw=red,line width=0.5mm] ($ (img) + (1.1,0.9) $) rectangle ++(0.6,0.6);
\node [image,right=of last,alias=rlast] (img){\includegraphics[width=\linewidth]{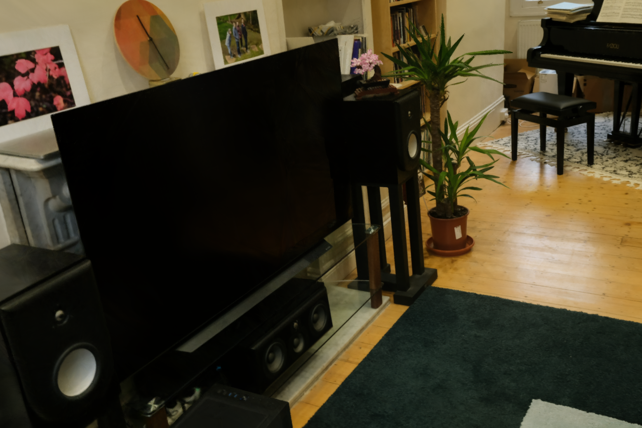}};
\draw[draw=red,line width=0.5mm] ($ (img) + (1.1,0.9) $) rectangle ++(0.6,0.6);
\node [image,right=of rlast] (img)           {\includegraphics[width=\linewidth]{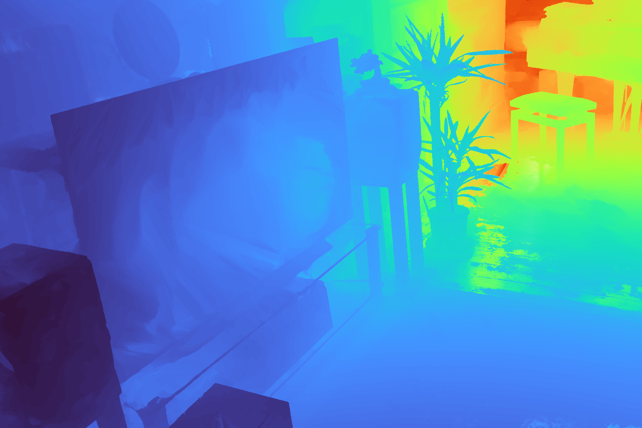}};
\draw[draw=red,line width=0.5mm] ($ (img) + (1.1,0.9) $) rectangle ++(0.6,0.6);
    
\node [label,above=-1pt of img00] {ground truth};
\node [label,above=-1pt of img01] {prediction};
\node [label,above=-1pt of img02] {depth};
\end{tikzpicture}

\caption{\textbf{Results on the indoor scenes from Mip-NeRF 360 dataset \cite{barron2022mipnerf360}.}
We show the \textbf{ground-truth} image, the \textbf{prediction}, and the \textbf{predicted depth map} on scenes: \textit{bonsai} \textbf{(top)}, \textit{counter}, \textit{kitchen}, and \textit{room} \textbf{(bottom)}.
The \textcolor{red}{red} squares highlight regions where there are visible errors in the images. 
\ours is able to represent even fine details, such as texts on products in the \textit{counter} scene, well.
\label{fig:supp-360-1}
}
\end{figure*}
\begin{figure*}[ht!]
\centering
\small
\begin{tikzpicture}[
 image/.style = {text width=0.31\linewidth, 
                 inner sep=0pt, outer sep=0pt},
label/.style = { minimum height=0.4cm },
node distance = 1pt and 1pt
                        ] 
\path coordinate(last);
\node [image,below=of last,alias=last] (img00) {\includegraphics[width=\linewidth]{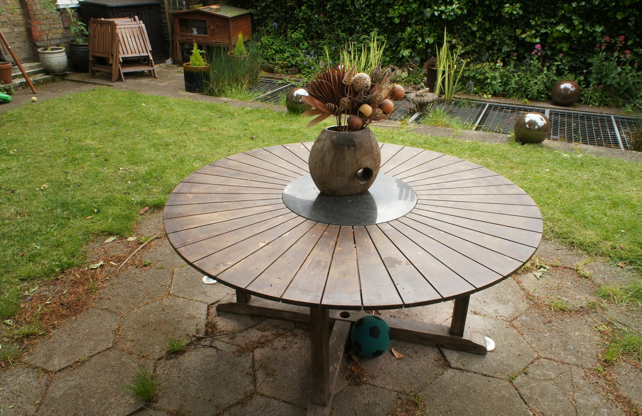}};
\draw[draw=red,line width=0.5mm] ($ (img00) + (0.2,-0.3) $) rectangle ++(0.6,0.6);
\node [image,right=of last,alias=rlast] (img01){\includegraphics[width=\linewidth]{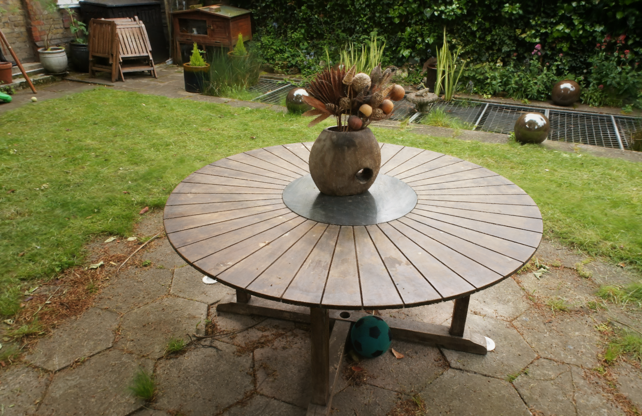}};
\draw[draw=red,line width=0.5mm] ($ (img01) + (0.2,-0.3) $) rectangle ++(0.6,0.6);
\node [image,right=of rlast] (img02)           {\includegraphics[width=\linewidth]{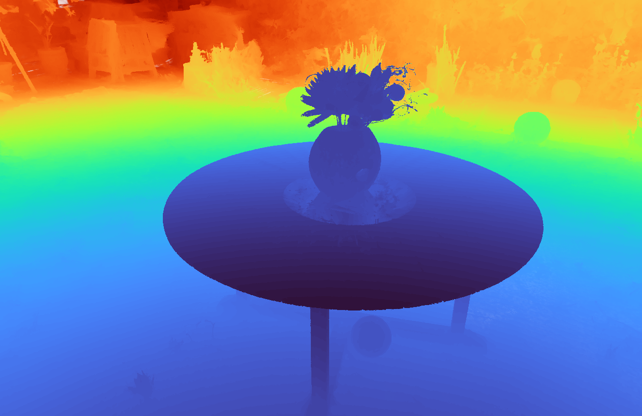}};
\draw[draw=red,line width=0.5mm] ($ (img02) + (0.2,-0.3) $) rectangle ++(0.6,0.6);

\node [image,below=of last,alias=last] (img) {\includegraphics[width=\linewidth]{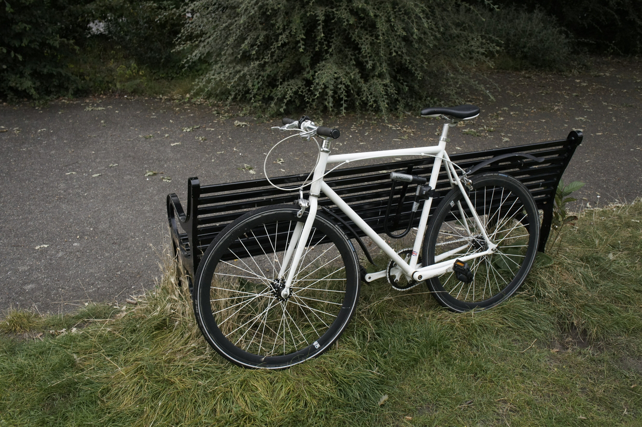}};
\draw[draw=red,line width=0.5mm] ($ (img) + (-2.5,-1.76) $) rectangle ++(0.6,0.6);
\node [image,right=of last,alias=rlast] (img){\includegraphics[width=\linewidth]{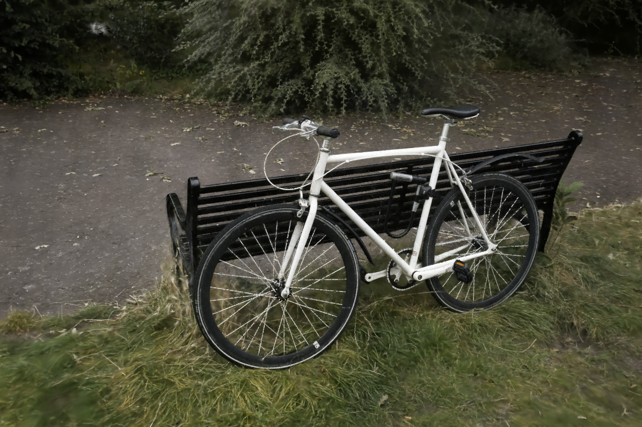}};
\draw[draw=red,line width=0.5mm] ($ (img) + (-2.5,-1.76) $) rectangle ++(0.6,0.6);
\node [image,right=of rlast] (img)           {\includegraphics[width=\linewidth]{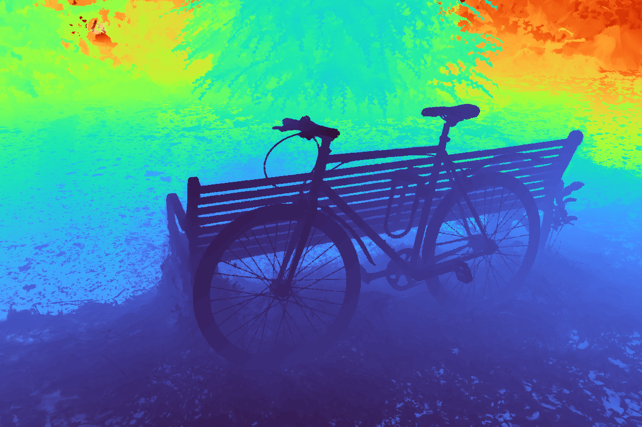}};
\draw[draw=red,line width=0.5mm] ($ (img) + (-2.5,-1.76) $) rectangle ++(0.6,0.6);

\node [image,below=of last,alias=last] (img) {\includegraphics[width=\linewidth]{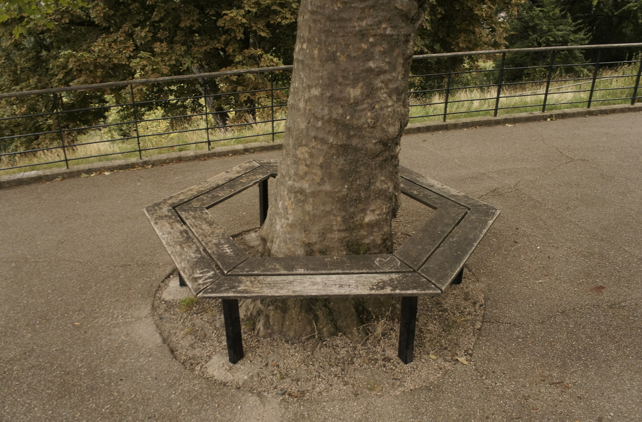}};
\draw[draw=red,line width=0.5mm] ($ (img) + (-2.5,0.6) $) rectangle ++(0.6,0.6);
\node [image,right=of last,alias=rlast] (img){\includegraphics[width=\linewidth]{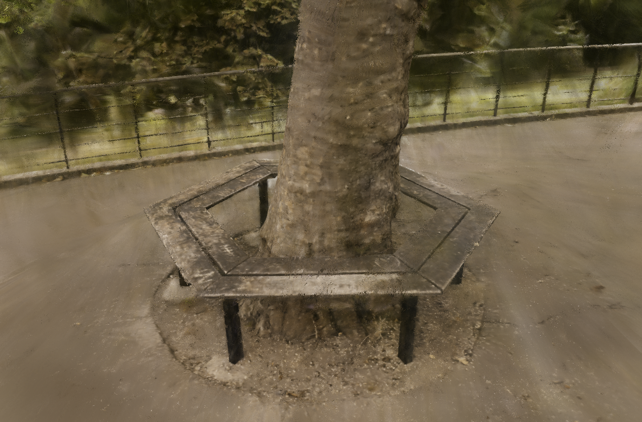}};
\draw[draw=red,line width=0.5mm] ($ (img) + (-2.5,0.6) $) rectangle ++(0.6,0.6);
\node [image,right=of rlast] (img)           {\includegraphics[width=\linewidth]{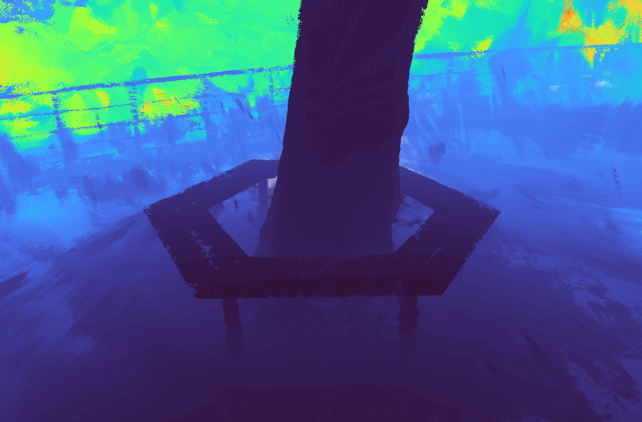}};
\draw[draw=red,line width=0.5mm] ($ (img) + (-2.5,0.6) $) rectangle ++(0.6,0.6);

\node [image,below=of last,alias=last] (img) {\includegraphics[width=\linewidth]{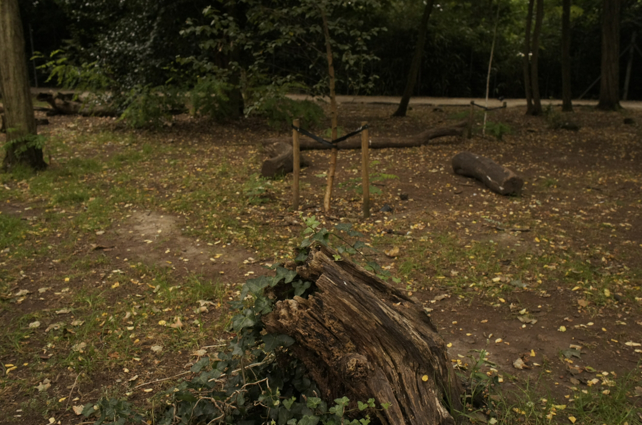}};
\draw[draw=red,line width=0.5mm] ($ (img) + (-2.5,-1.0) $) rectangle ++(0.6,0.6);
\node [image,right=of last,alias=rlast] (img){\includegraphics[width=\linewidth]{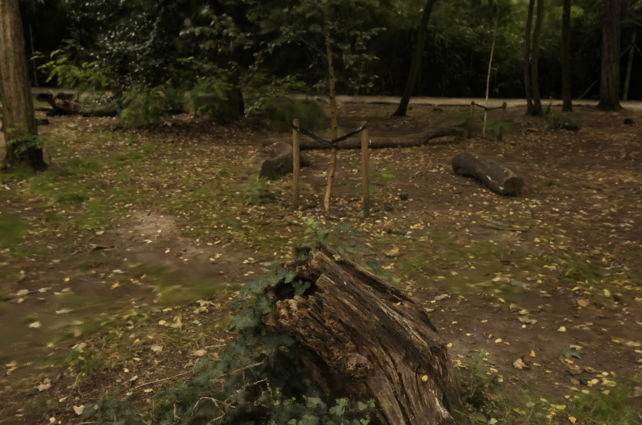}};
\draw[draw=red,line width=0.5mm] ($ (img) + (-2.5,-1.0) $) rectangle ++(0.6,0.6);
\node [image,right=of rlast] (img)           {\includegraphics[width=\linewidth]{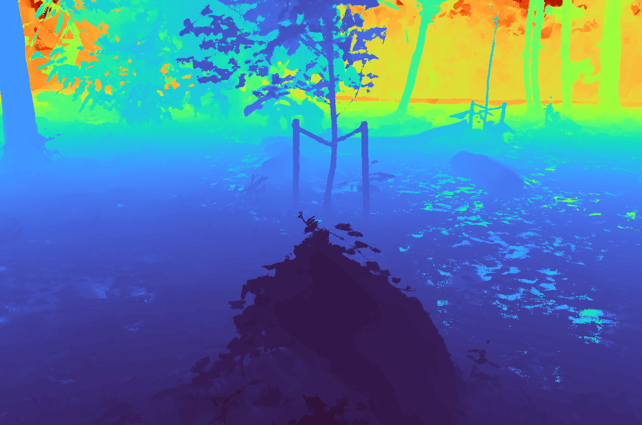}};
\draw[draw=red,line width=0.5mm] ($ (img) + (-2.5,-1.0) $) rectangle ++(0.6,0.6);

\node [image,below=of last,alias=last] (img) {\includegraphics[width=\linewidth]{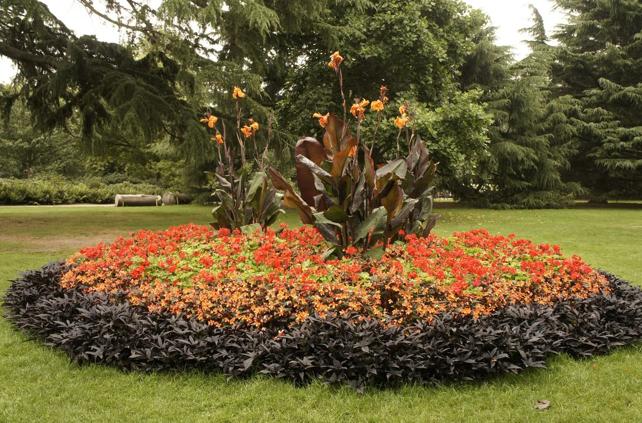}};
\draw[draw=red,line width=0.5mm] ($ (img) + (-2.5,-1.76) $) rectangle ++(0.6,0.6);
\node [image,right=of last,alias=rlast] (img){\includegraphics[width=\linewidth]{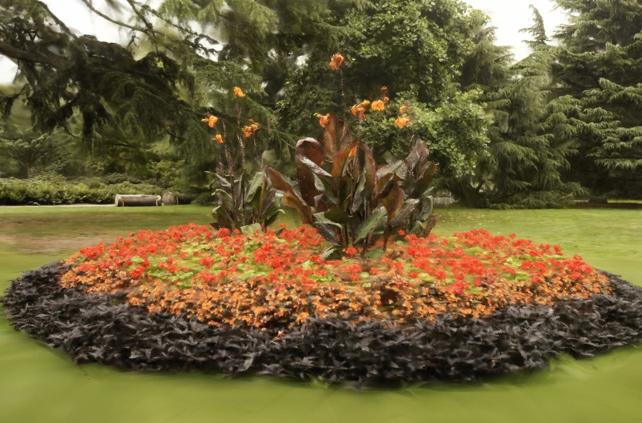}};
\draw[draw=red,line width=0.5mm] ($ (img) + (-2.5,-1.76) $) rectangle ++(0.6,0.6);
\node [image,right=of rlast] (img)           {\includegraphics[width=\linewidth]{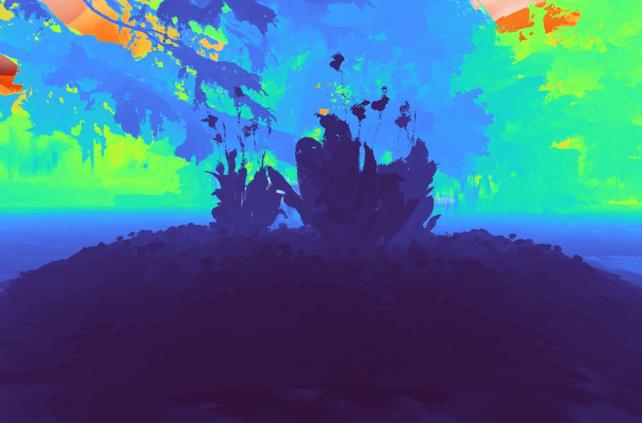}};
\draw[draw=red,line width=0.5mm] ($ (img) + (-2.5,-1.76) $) rectangle ++(0.6,0.6);

\node [label,above=-1pt of img00] {ground truth};
\node [label,above=-1pt of img01] {prediction};
\node [label,above=-1pt of img02] {depth};
\end{tikzpicture}

\caption{\textbf{Results on the outdoor scenes from Mip-NeRF 360 dataset \cite{barron2022mipnerf360}.}
We show the \textbf{ground-truth} image, the \textbf{prediction}, and the \textbf{predicted depth map} on scenes: \textit{garden} \textbf{(top)}, \textit{bicycle}, \textit{treehill}, \textit{stump}, and \textit{flowers} \textbf{(bottom)}.
 The \textcolor{red}{red} squares highlight regions where there are visible errors in the images. 
\ours is not able to represent the grass and the ground in \textit{flowers} and \textit{treehill} scenes well, and we can see blur artefacts. On the \textit{garden} scene, the reconstruction achieves high fidelity except for the centre of the table, where the reflections are slightly incorrect.
\label{fig:supp-360-2}
}
\end{figure*}
We also extend the results on the Mip-NeRF 360 dataset \cite{barron2022mipnerf360}, presented in Table~8 and Figure~4 in the main paper, by showing detailed results for each scene. We follow the same training and evaluation procedure as Mip-NeRF 360~\cite{barron2022mipnerf360}, and for the outdoor and indoor scenes, we train and evaluate with 4x and 2x downsampled images, respectively.
The PSNR, SSIM, and LPIPS (Alex) \cite{zhang2018lpips} results are presented in Table~\ref{tab:supp-mipnerf360}.
In terms of SSIM, our approach performs slightly worse %
than Point-Based Neural Rendering~\cite{kopanas2021point}, Stable View Synthesis~\cite{riegler2021stable}, and Mip-NeRF 360~\cite{barron2022mipnerf360}. 
In terms of the PSNR and LPIPS metrics, our approach performs comparable or better than these baselines. 
\Eg, 
the state-of-the-art Mip-NeRF 360~\cite{barron2022mipnerf360} %
has a slightly better PSNR and \ours has a slightly better LPIPS. 
We %
typically outperform Stable View Synthesis \cite{riegler2021stable} and competitors other than Mip-NeRF 360 in terms of PSNR. Note that similarly to \ours, Stable View Synthesis uses a geometric prior as input -- a mesh instead of a point cloud.

We also show qualitative results for indoor and outdoor Mip-NeRF 360 scenes in Figures~\ref{fig:supp-360-1} and \ref{fig:supp-360-2}. %
We notice more artefacts in the outdoor scenes compared to indoor ones. Small, high-frequency details such as grass are not represented well, which can be visible, \eg, in the \textit{flowers} scene. 
One potential cause for this behaviour, which we plan to investigate in future work,  could be that the poses estimated outdoors (where the camera is typically farther away from the scene than indoors) are noisier which makes it harder to recover fine details. %
For the indoor scenes, our model is able to represent the scenes with very high fidelity,  %
including very fine details such as texts on products in the \textit{counter} scene.

\clearpage
\section{Varying the number of input points}\label{sec:ablation-num-points}
\definecolor{plt-blue}{HTML}{1f77b4}
\definecolor{plt-orange}{HTML}{ff7f0e}
\definecolor{plt-green}{HTML}{2ca02c}
\definecolor{plt-red}{HTML}{d62728}
\begin{figure*}[ht!]
\subfloat[\centering PSNR]{{
    \includegraphics[width=0.32\columnwidth]{assets/ablations/num-points-ablation-multi-psnr}
}}        
\subfloat[\centering SSIM]{{
\includegraphics[width=0.32\columnwidth]{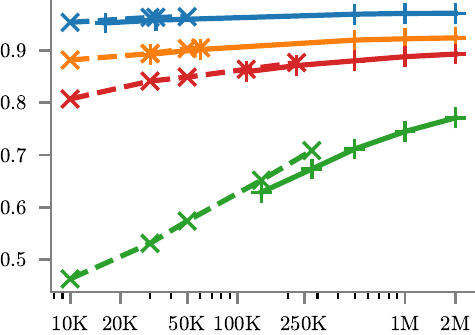}
}}
\subfloat[\centering LPIPS]{{
\includegraphics[width=0.32\columnwidth]{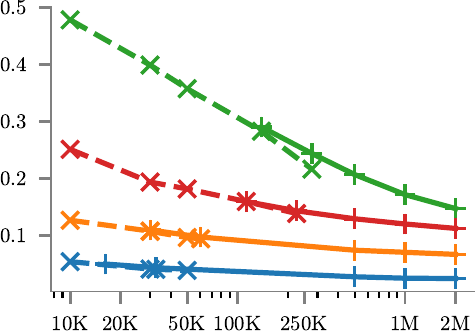}
}}
\vspace{0.2\baselineskip}
\caption{\textbf{Performance with different sizes and quality of the input point cloud.}
\textbf{a, b, c)} shows the PSNR, SSIM, and LPIPS \cite{zhang2018lpips} with different sizes of input point clouds (PSNR is repeated from the main paper). The solid and dashed lines represent the dense and coarse COLMAP reconstructions. The following scenes were evaluated (best to worst): \textcolor{plt-blue}{\textit{tt/family}}, \textcolor{plt-red}{\textit{360/room}}, \textcolor{plt-orange}{\textit{tt/truck}}, and \textcolor{plt-green}{\textit{360/garden}} from the Tanks and Temples~\cite{knapitsch2017tanks} (tt) and the Mip-NeRF 360~\cite{barron2022mipnerf360} (360-indoor/360-outdoor) datasets. As expected, the quality increases with the number of points, but also sparse reconstruction performs better at the same number of points.
\label{fig:ablation-num-input-points}
}
\end{figure*}
We conducted a study on the effect of different sizes of the input point cloud.
In the main paper, we presented the PSNR results on four scenes. Here we also show the SSIM and LPIPS.
We consider both the sparse and dense COLMAP reconstructions and analyse the performance as we sub-sample the points or add more points randomly. In order to save computational resources, we only train the method for $100$k iterations.
The results are visualised in Figure~\ref{fig:ablation-num-input-points}.

As expected, the performance improves with the number of points used, as it leads to a finer subdivision of the scene around the surface.
Note how the performance of dense reconstruction is lower than sparse reconstruction with the same size. Also, note how the performance of \textit{360/garden} increases when adding randomly sampled points to the sparse point cloud. 
For \ours it is important to have a dense-enough coverage in regions close to surfaces and randomly subsampling a larger point cloud may miss some regions. Similarly, adding more points at random in proximity to existing ones helps as it increases the density in those regions.
For all experiments in the paper and the \suppmat, we used random sub-sampling. 
An interesting direction for future work is to investigate more sophisticated strategies that, \eg, sample more points around fine details such as corners and edges and fewer points in planar regions.

}

\twocolumn
{\small
\bibliographystyle{ieee_fullname}
\bibliography{egbib}
}
\end{document}